\documentclass[final,12pt]{clear2027_preprint}

\usepackage{times}

\clearauthor{%
 \Name{Jacqueline Maasch} \Email{jam887@cornell.edu}\\
 \addr Cornell Tech
 \AND
 \Name{Shu Wan} \Email{shu.wan@abel.ai}\\
 \addr Abel AI 
 \AND
 \Name{Hank Zhu} \Email{hank.zhu@abel.ai}\\
 \addr Abel AI 
 \AND
 \Name{Huan Liu} \Email{}\\
 \addr Arizona State University
 \AND
 \Name{Zhiguang Wang} \Email{}\\
 \addr Abel AI %
}

\usepackage[]{natbib}

\usepackage{wrapfig}
\usepackage{setspace}

\usepackage{amsmath}
\usepackage{amssymb}
\usepackage{mathtools}
\usepackage{dsfont}
\usepackage{mathrsfs}

\newcommand{\ind}{\perp\!\!\!\perp} 
\newcommand{\nind}{\not\!\perp\!\!\!\perp} 

\newcommand{\g}{\mathcal{G}}
\newcommand{\z}{\mathbf{Z}}
\newcommand{\pa}{\mathrm{pa}}
\newcommand{\ch}{\mathrm{ch}}
\newcommand{\an}{\mathrm{an}}
\newcommand{\de}{\mathrm{de}}
\newcommand{\nd}{\mathrm{nd}}

\usepackage{xspace}

\newcommand{\cit}{\texttt{CIT}\xspace}

\usepackage[dvipsnames]{xcolor}
\usepackage{hyperref}
\definecolor{glaucous}{rgb}{0.38, 0.51, 0.71}
\hypersetup{
    colorlinks=true, 
    linkcolor=glaucous,
    filecolor=Blue,      
    urlcolor=Blue,
    citecolor=Blue 
    }

\usepackage[toc,page,header]{appendix}
\usepackage{minitoc}
\renewcommand \thepart{}
\renewcommand \partname{}

\usepackage{pifont} 
\definecolor{chestnut}{rgb}{0.8, 0.36, 0.36}

\usepackage{algorithm}
\usepackage[noend]{algorithmic}
\newcommand{\algorithmicinput}{\textbf{Input:}}
\newcommand{\algorithmicoutput}{\textbf{Output:}}
\newcommand{\INPUT}{\item[\algorithmicinput]}
\newcommand{\OUTPUT}{\item[\algorithmicoutput]}
\newcommand{\algorithmicassumptions}{\textbf{Assumptions:}}
\newcommand{\ASSUMPTIONS}{\item[\algorithmicassumptions]}
\newcommand{\algorithmicnotation}{\textbf{Notation:}}
\newcommand{\NOTATION}{\item[\algorithmicnotation]}

\usepackage{tikz}
\usetikzlibrary {arrows.meta,arrows,patterns,decorations.pathreplacing,chains,positioning,shapes,shapes.geometric,calc,backgrounds,automata,fit}
\usepackage{pgfplots}
\pgfplotsset{compat=1.15}
\usepackage{pifont}
\usepackage{tikz-cd}

\pgfplotsset{
    resaxis/.style={tick label style={font=\footnotesize}, label style={font=\small}, title style={font=\small},
        axis lines=left, axis line style={gray!75}, ymajorgrids, grid style={gray!20}, legend cell align=left,
        legend style={font=\footnotesize, draw=none, fill=none, /tikz/every even column/.append style={column sep=7pt}}},
    mITPD/.style={parents, very thick, mark=*, mark size=2.2pt, mark options={solid}},
    mLasso/.style={black!60, thick, mark=diamond*, mark size=2.6pt, mark options={solid}},
    mDFPC/.style={black!60, thick, mark=square*, mark size=1.7pt, mark options={solid}},
    mIPaDL/.style={black!60, thick, mark=triangle*, mark size=2.3pt, mark options={solid}},
    mIAMB/.style={black!35, thick, mark=pentagon*, mark size=2.1pt, mark options={solid}},
    mFullCI/.style={black!35, thick, mark=x, mark size=2.6pt, mark options={solid}},
    mPCMCI/.style={glaucous, thick, mark=*, mark size=1.9pt, mark options={solid}},
    mDYNO/.style={glaucous, thick, mark=square*, mark size=1.7pt, mark options={solid}},
    mRef/.style={black!60, thick, only marks, mark=o, mark size=2.1pt}
}

\usepackage{enumitem}
\setlist[itemize]{noitemsep, topsep=0pt, leftmargin=*}
\setlist[enumerate]{noitemsep, topsep=0pt, leftmargin=*}

\usepackage{colortbl} 
\usepackage{array} 
\usepackage{adjustbox}  
\usepackage{multirow} 
\usepackage{booktabs} 
\usepackage{tabularray}
\UseTblrLibrary{booktabs}

\definecolor{X_color}{rgb}{0.96, 0.96, 0.96}
\definecolor{Y_color}{rgb}{0.96, 0.96, 0.96}
\definecolor{Z1_color}{rgb}{0.63, 0.79, 0.95}
\definecolor{Z2_color}{rgb}{1.0, 0.74, 0.85}
\definecolor{Z4_color}{rgb}{0.6, 0.8, 0.2}
\definecolor{Z5_color}{rgb}{1.0, 0.75, 0.0} 
\definecolor{Z6_color}{rgb}{0.54, 0.17, 0.89}
\definecolor{Z7_color}{rgb}{0.91, 0.67, 0.81}
\definecolor{Z8_color}{rgb}{0.0, 0.65, 0.58}
\definecolor{parents}{rgb}{0.81, 0.09, 0.13}

\doparttoc 
\faketableofcontents 

\newcommand{\mytitle}{Time Series Structure Learning with Local Causal Parent Discovery}

\title[\mytitle]{\mytitle}

\begin{document}

\maketitle

\begin{abstract}
    Learning and inference on time series can pose significant statistical challenges, including autocorrelated observations, distribution shift over time, and the curse of dimensionality. 
    These challenges complicate the practical application of causal discovery algorithms, which already suffer from high computational and sample complexity in the general case. To mitigate these issues, this work introduces \underline{I}terative \underline{T}emporal \underline{P}arent \underline{D}iscovery (ITPD): a sound and complete constraint-based procedure for causal structure learning on multivariate time series with causal nonstationarity. ITPD takes a local-to-global learning approach that discovers the global graph by sequentially discovering the parent set for each variable. ITPD exploits temporal structural priors, graph partitioning, and the local Markov condition to swap higher-order conditional independence tests for lower-order alternatives. With an exact oracle, ITPD performed the least number of tests and used the smallest average conditioning set sizes relative to baselines. In benchmarking simulations, ITPD sat on the Pareto frontier with respect to F1 and wall-clock time and was faster than baselines on large graphs ($\geq$ 1000 nodes). For graphs with 2000 nodes, ITPD ran $2.3\times$, $9.1\times$, and $40.9\times$ faster than iterative IAMB, DF-PC, and PCMCI, respectively. 
\end{abstract}

\begin{keywords}%
  Time series, causal discovery, graphical models
\end{keywords}


\section{Introduction}

Temporal data naturally arise in many high-impact domains, including biomedicine, signal processing, meteorology, ecology, finance, prediction markets, and dynamical systems research. In these settings, knowledge of how causal relationships recur and change over time can support decision-making, forecasting, and intervention. In practice, temporal causal structure is rarely known a priori and must be learned from data \citep{glymour2019review}. However, many advancements in statistics and machine learning assume \textit{independent and identically distributed} (iid) random variables and some form of \textit{stationarity} \citep{hastie2009elements}, complicating their direct application to temporal data with autocorrelation and distribution shift (Figure \ref{fig:window_v_full_time}). Such challenges motivate the design of special methods for time series analysis \citep{brockwell2009time}, including temporal causal discovery algorithms \citep{assaad2022survey,runge2023causal, hasan2023survey}.

This work considers the problem of temporal causal discovery under distribution shift. We restrict our attention to the constraint-based paradigm, which uses sequential conditional independence (CI) tests to infer graphical structure \citep{spirtes2001causation}. This paradigm benefits from a wealth of theoretical results, minimal innate parametric assumptions, and explainability. However, constraint-based discovery suffers from several drawbacks in the general case, including high computational complexity and high sample complexity \citep{shiragur2024causal}. 

\definecolor{A_color}{HTML}{FDD5B1}
\definecolor{B_color}{HTML}{CCCCFF}
\definecolor{C_color}{HTML}{FDDDE6}

\begin{wrapfigure}[16]{r}{0.50\linewidth}
    \centering
\tikzset{
  A/.style={
    circle,
    draw=black,
    fill=A_color!30,
    thick,
    inner sep=3pt,
  },
  B/.style={
    circle,
    draw=black,
    fill=B_color!30,
    thick,
    inner sep=3pt,
  },
  C/.style={
    circle,
    draw=black,
    fill=C_color!30,
    thick,
    inner sep=3pt,
  }
}

\begin{tikzpicture}[
every edge/.append style={thick, black, >={Latex[length=1.75mm,width=1.75mm]}},
scale=0.8,
font=\footnotesize
]


\node[A] (A1) at (0,4) {A$_1$};
\node[A] (A2) at (1.5,4) {A$_2$};
\node[A] (A3) at (3,4) {A$_3$};
\node[A] (A4) at (4.5,4) {A$_4$};
\node[A] (A5) at (6,4) {A$_5$};
\node[A] (A6) at (7.5,4) {A$_6$};
\node[] (A7) at (8.25,4) {$\cdots$?};

\node[B] (B1) at (0,2.8) {B$_1$};
\node[B] (B2) at (1.5,2.8) {B$_2$};
\node[B] (B3) at (3,2.8) {B$_3$};
\node[B] (B4) at (4.5,2.8) {B$_4$};
\node[B] (B5) at (6,2.8) {B$_5$};
\node[B] (B6) at (7.5,2.8) {B$_6$};
\node[] (B7) at (8.25,2.8) {$\cdots$?};

\node[C] (C1) at (0,1.6) {C$_1$};
\node[C] (C2) at (1.5,1.6) {C$_2$};
\node[C] (C3) at (3,1.6) {C$_3$};
\node[C] (C4) at (4.5,1.6) {C$_4$};
\node[C] (C5) at (6,1.6) {C$_5$};
\node[C] (C6) at (7.5,1.6) {C$_6$};
\node[] (C7) at (8.25,1.6) {$\cdots$?};

\path (A1) edge[->,gray!50] (A2);
\path (A2) edge[->,gray!50] (A3);
\path (A3) edge[->,gray!50] (A4);
\path (A4) edge[->,gray!50] (A5);
\path (A5) edge[->,gray!50] (A6);

\path (B1) edge[->,gray!50] (B2);
\path (B2) edge[->,gray!50] (B3);
\path (B3) edge[->,gray!50] (B4);
\path (B4) edge[->,gray!50] (B5);
\path (B5) edge[->,gray!50] (B6);

\path (C1) edge[->,gray!50] (C2);
\path (C2) edge[->,gray!50] (C3);
\path (C3) edge[->,gray!50] (C4);
\path (C4) edge[->,gray!50] (C5);
\path (C5) edge[->,gray!50] (C6);

\path (A1) edge[->] (B2);
\path (C2) edge[->] (B3);
\path (B4) edge[->] (A5);
\path (B4) edge[->] (C5);
\path (C5) edge[->] (B6);

\node[draw,thick,dotted,fit=(A1) (B1) (C1) (A2) (B2) (C2) (A3) (B3) (C3)] {};
\node[] (label) at (1.5,0.5) {\footnotesize\textit{Window with max lag = 2}};

\end{tikzpicture}

    \caption{Full time graph $\g$ with autocorrelation and causal nonstationarity. Under the false assumption that $\g$ can be represented by a stationary window graph with a maximum lag of 2, values at time $t > 3$ would be incorrectly forecasted. 
    }
    \label{fig:window_v_full_time}
\end{wrapfigure}
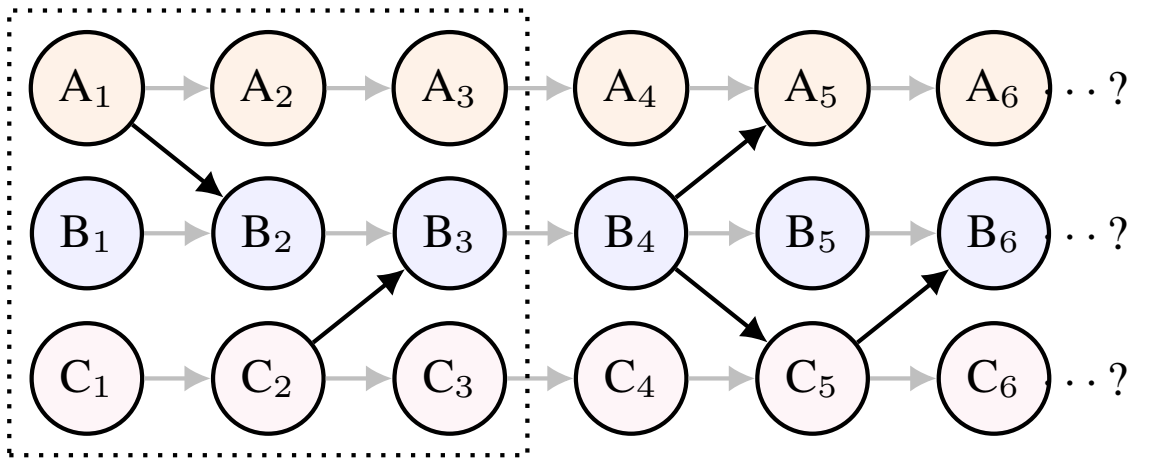

The computational complexity of constraint-based discovery is often measured with respect to (wrt) the worst-case number of CI tests required \citep{spirtes2001causation}. For the  general case, this is exponential wrt some graph parameter (e.g., total nodes, maximum in-degree, etc.). Unless assumptions are placed on the data generating process, it has been shown that the number of CI tests is lower bounded by a value that is exponential in the maximum undirected clique size of the essential graph \citep{mones2026number}. CI testing is known to be a hard problem in the general case \citep{shah2020hardness}, with exponential sample complexity wrt the size of the conditioning set. Exponential time and sample complexity can prevent tractable discovery on large data, motivating diverse strategies for pruning CI tests and reducing their \textit{order} (i.e., the number of variables in the conditioning set) \citep{shiragur2024causal, kim2026deduce}. We extend this effort to the time series setting, where high-dimensional data often arise.

The search space of time series directed acyclic graphs (DAGs) can be significantly reduced by leveraging known problem structure, such as temporal structural priors or other graph parameters. Exploiting structural priors for efficiency gains is a common strategy in \textit{local causal discovery} algorithms that only learn contextually relevant subgraphs \citep{maasch2024local}. This work applies lessons from local structure learning to time series whose causal graphs are nonstationary. As an inspiration point, we naively apply previously introduced local discovery algorithms \citep{tsamardinos2003algorithms,maasch2025local} in iterative fashion to obtain the global graph. As these local discovery algorithms were not intended for global time series discovery, this naive approach can entail extraneous CI tests or unnecessarily large conditioning sets. To mitigate this, we introduce an iterative procedure that is tailored for nonstationary temporal graphs. 

\paragraph{Contributions} Our methodological and empirical contributions are as follows.
\begin{enumerate}[leftmargin=0.9cm]
    \item [\S\ref{sec:method}] \textit{Local-to-global temporal causal discovery algorithms.} 
        \begin{itemize}[leftmargin=0.8cm]
            \item [\S\ref{sec:ipadl_in_text}] \textit{Iterative Parent Discovery on Leaves (IPaDL)}. We present Algorithm \ref{alg:itpd_naive}, a sound and complete naive baseline that extends a previously introduced local parent discovery algorithm to the nonstationary time series setting. We use the inefficiencies of this simplistic adaptation as an inspiration and departure point for our core algorithmic contribution. 
            \item [\S\ref{sec:itpd}] \textit{Iterative Temporal Parent Discovery (ITPD).} We present Algorithm \ref{alg:itpd}, a sound and complete causal discovery algorithm for nonstationary time series designed to reduce reliance on high-order CI tests. Additionally, we provide support for manually-derived theoretical results via AI-assisted autoformalization in Lean.
        \end{itemize}
    \item [\S\ref{sec:empirics}] \textit{Empirical validation and open code.} Theoretical claims are numerically verified on synthetic and semi-synthetic data. ITPD sat on the Pareto frontier wrt F1 and wall-clock time as well as F1 and unique CI tests per variable. On graphs with 1000 to 2000 variables, ITPD was the fastest running method tested. Code is available on GitHub.\footnote{\href{https://anonymous.4open.science/r/itpd-0FC0/README.md}{\texttt{https://anonymous.4open.science/r/itpd-0FC0}}} 
    \item [\S\ref{appendix:ablations}] \textit{Negative results post-mortem}. We provide a supplementary discussion of a systematic ablation study, highlighting strategies that failed to contribute to improved efficiency in ITPD. 
\end{enumerate}

\section{Related Works}

\paragraph{Time Series Causal Discovery} Diverse strategies have been proposed for learning temporal structure from data. An historically dominant approach has been Granger causality and its variants \citep{granger1969investigating,geweke1982measurement}, including neural Granger causality \citep{marcinkevivcs2021interpretable,tank2022neural,lowe2022amortized}. Constraint-based algorithms have been extended to this setting, including FCI \citep{entner2010causal}. Most notably,  the PCMCI family of algorithms extends PC to causally stationary settings under causal sufficiency and latent variables, with and without contemporaneous links \citep{runge2019detecting,pmlr-v124-runge20a,gunther2023causal,runge2023causal}. Methods that assume functional causal models \citep{fujiwara2023causal} and  neural methods also abound \citep{pamfil2020dynotears,sun2023nts}. 
See \citet{assaad2022survey,hasan2023survey} for thorough surveys on time series causal discovery.

\paragraph{Local Causal Discovery} Often, the problem setting for causal discovery does not necessitate knowledge of the full graphical structure. Instead, learning \textit{local substructures} is often sufficient to answer the scientific questions of interest. This emphasis on tailored local discovery can yield more computationally and sample efficient search procedures. Hence, \textit{local causal discovery} \citep{cooper1997simple} is a popular strategy for many problems, including parent and child discovery \citep{gao2015local}, Markov blanket (MB) discovery \citep{tsamardinos2003time,tsamardinos2003algorithms,aliferis2010local,wang_discovering_2014}, valid adjustment set discovery \citep{cheng_local_2022,gupta_local_2023,maasch2024local,maasch2025local,li2026local}, identifying necessary and sufficient explanatory variables \citep{cai2023learning}, and causal fairness analysis \citep{maasch2025local}. Local learning can also be used as a subroutine for global causal discovery in static DAGs \citep{hiremath2024hybrid} and in the time series setting \citep{fujiwara2023causal}, as we do in the present work.



\section{Preliminaries}

\paragraph{Notation} We represent random variables by uppercase letters and their realizations by lowercase, with sets of variables and vector-values in bold (e.g., $V = v$, $\mathbf{V} = \mathbf{v}$). Graphs are denoted by calligraphic script (e.g., $\mathcal{G}$). Given a variable $V$ in DAG $\mathcal{G}$, we denote its set of parents, children, ancestors, descendants,  nondescendants, and spouses in $\mathcal{G}$ as $\pa(V)$, $\ch(V)$, $\an(V)$, $\de(V)$, and $\nd(V)$, respectively. 
Let the ancestor set of a set be denoted $\an(\mathbf{S}) \coloneqq \bigcup_{W\in\mathbf S}\an(W)$. Sets $\widehat{\an}$($X$) and $\widehat{\pa}$($X$) denote inferred ancestor and parent sets for $X$, respectively. Assume notationally that $\pa(X) \subseteq \an(X)$ and $\ch(X) \subseteq \de(X)$. We denote the probabilistic independence of random variables by $\ind$ and the independence of nodes in $\g$ ($d$-separation) by $\ind_\g$. Let $X \to Y$ denote that variable $X$ is a direct cause of another variable $Y$, while $X \to \cdots \to Y$ denotes an indirect causal path. The absence of a directed edge between $X$ and $Y$ is denoted $X \not\to Y$.

\subsubsection{Temporal Causal Models} We consider structural causal models (SCMs) in the nonparametric structural equation models with independent errors (NPSEM-IE) regime \citep{wang2025causal}. 

\begin{definition}[Structural causal model (SCM),  \citealt{bareinboim2022pch}] \label{def:scm}
    An SCM is a tuple $\mathcal{M} \coloneqq \langle \mathbf{U}, p(\mathbf{u}), \mathbf{V}, \mathcal{F} \rangle$ where $\mathbf{U}$ is the set of exogenous variables explained by  mechanisms external to $\mathcal{M}$, $p(\mathbf{u})$ is the distribution over $\mathbf{U}$, $\mathbf{V}$ is the set of endogenous variables explained by $\mathbf{U} \cup \mathbf{V}$, and $\mathcal{F}$ is the set of structural functions such that $v^i = f^i(\pa(v^i), u^i)$ for endogenous parents $\pa(v^i)$ and exogenous context $u^i$.
\end{definition}
We restrict our attention to  SCMs whose graphical representations are DAGs.  Under the NPSEM-IE, SCM $\mathcal{M}$ features jointly independent noise terms: $U^i \ind U^j$ for all distinct $i,j$. Further, we take SCM $\mathcal{M}$ to represent a discrete-time stochastic process. 

\begin{definition}[Discrete-time stochastic process]
\label{def:stochastic_process}
    Let stochastic process $\mathbf{V} \in \mathbb{R}^{T \times N}$ be a set of random variables that are time-indexed over a finite horizon by $t \in [1,\dots,T]$. 
    Let $\mathbf{V}_t \coloneqq \{V^n_t\}_{n=1}^{N}$ denote the $N$ variables at a given time step. Let $\mathbf{V}^n \coloneqq \{V^n_t\}_{t=1}^T$ denote the full time series for the $n^{th}$ variable.
\end{definition}
As we take endogenous variables $\mathbf{V}$ to be a stochastic process, we adopt a temporal NPSEM-IE termed the \textit{time series model with independent noise} (TiMINo) \citep{peters2013causal}. The local Markov condition is known to hold under TiMINo \citep{peters2013causal}. Following from local Markov we obtain the global Markov condition. 

\begin{definition}[Local Markov condition 
]
\label{def:local_markov}
    A distribution $p$ satisfies the local Markov condition with respect to DAG $\g = (\mathbf{V}, \mathbf{E})$ if every variable $V \in \mathbf{V}$ is conditionally independent of its (nonparent) nondescendants given its parents in $\g$:
    \begin{align}
        V \ind \nd(V) \setminus \pa(V) \mid \pa(V).
    \end{align}
\end{definition}



\begin{definition}[Global Markov condition]
    A distribution $p$ satisfies the global Markov condition with respect to DAG $\g = (\mathbf{V}, \mathbf{E})$ if for all disjoint sets $\mathbf{X}, \mathbf{Y}, \mathbf{Z} \subset \mathbf{V}$,
    \begin{align}
        \mathbf{X} \ind_\g \mathbf{Y} \mid \mathbf{Z} \Longrightarrow \mathbf{X} \ind \mathbf{Y} \mid \mathbf{Z}.
    \end{align}
\end{definition}

\begin{definition}[Faithfulness, \citealt{uhler2013geometry}]
    Let $\mathbf{V}$ be a set of random variables that can be graphically represented as a DAG $\g$. Distribution $p$ is faithful to $\g$ when all probabilistic independencies present in $p$ are exactly those graphical separations (\textit{d}-separations) entailed by the local Markov condition applied to  $\g$:
    \begin{align}
        \mathbf{X} \ind \mathbf{Y} \mid \mathbf{Z} \Longrightarrow \mathbf{X} \ind_\g \mathbf{Y} \mid \mathbf{Z}.
    \end{align}
\end{definition}
Following from global Markov and faithfulness, for  distribution $p(\mathbf{V})$ with DAG $\g = (\mathbf{V}, \mathbf{E})$ and disjoint $\mathbf{X},\mathbf{Y}, \mathbf{Z} \subseteq \mathbf{V}$, we assume $\mathbf{X} \ind \mathbf{Y} \mid \mathbf{Z} \iff \mathbf{X} \ind_\g \mathbf{Y} \mid \mathbf{Z}$.


Unlike PCMCI, this work does not require \textit{causal stationarity} (i.e., \textit{consistency throughout time} or structural stationarity), which assumes that any link $V^i_{t-\tau} \to V^j_t$ with lag $\tau$ recurs for all time steps $t$ \citep{runge2018causal}. Optionally, we can assume stationarity with a maximum lag of $\tau_{\max}$. The graphical structure representing all of $\mathbf{V}$ is known as the \textit{full time graph}, while the summary graph for a causally stationary stochastic process with lag $\tau_{\max}$ is known as a \textit{window graph} (Figure \ref{fig:window_v_full_time}).

\begin{definition}[Full time graph, 
\citealt{assaad2022survey}]
\label{def:full_time_graph}
    Let $\mathcal{G} = (\mathbf{V}, \mathbf{E})$ be a full time graph for discrete-time stochastic process $\mathbf{V}$. Nodes are $V^i_t \in \mathbf{V}$ for time steps $t \in [1, \dots, \infty]$. Edges $\mathbf{E}$ are lagged directed links $V^i_{t-\tau} \to V^j_t$ with a time lag of $\tau > 0$ for $i=j$ and $\tau \geq 0$ for $i \neq j$.
\end{definition}
As this work assumes no contemporaneous links, $\tau$ must be strictly greater than zero for $i \neq j$. While a full time graph is theoretically infinite, in practice we can visualize this as a finite graph on finite real-world measurements from $t = [1, \dots, T]$.

\begin{definition}[Window graph, \citealt{assaad2022survey}]
    Let $\mathcal{G} = (\mathbf{V}, \mathbf{E})$ be a window graph for discrete-time stochastic process $\mathbf{V}$ with a maximum causal lag of $\tau_{\max}$. Nodes are $\{V^n_t\}_{n=1}^N$ for time steps $t \in [1, \dots, \tau_{\max} + 1]$. Edges $\mathbf{E}$ are lag-specific directed links $V^i_{t-k} \to V^j_t$ for $0 \leq k \leq \tau_{\max}$ for $i \neq j$ and $0 < k \leq \tau_{\max}$ for $i = j$. 
    For a window graph with no contemporaneous links, $k$ must be strictly greater than zero for $i \neq j$.
\end{definition}

\setlength{\intextsep}{0pt}%

\begin{wraptable}[10]{r}{0.6\linewidth}
    \centering
    \vspace{-1cm}
    \begin{adjustbox}{max width=0.55\textwidth}
    \begin{tabular}{p{0.25cm} p{8.7cm}}
    \toprule
        $\z_1$ &  Confounders and their proxies. \\
        $\z_2$ &  Colliders and their proxies.\\
        $\z_3$ &  Mediators and their proxies. \\
        $\z_4$ &  $\nd(Y)$ where $\z_4 \ind X$ and $\z_4 \nind X \mid Y$. \\
        $\z_5$ &  Instruments and their proxies. \\
        $\z_6$ &  $\de(Y)$ where active paths with $X$ are mediated by $Y$. \\
        $\z_7$ &  $\de(X)$ where active paths with $Y$ are mediated by $X$. \\
        $\z_8$ &  Nodes that share no active paths with $X$ nor $Y$. \\
    \bottomrule
    \end{tabular}
    \end{adjustbox}
    \caption{Disjoint causal partitions wrt cause-effect pair $\{X,Y\}$. See \citet{maasch2024local} for formal definitions of partitions and their proxies.}
    \label{tab:partitions}
\end{wraptable}

\subsubsection{Causal Partitions} 
This work leverages the observation that any cause-effect pair $X \to \cdots \to Y$ in $\mathbf{V}$ induces a unique \textit{causal partitioning} of $\mathbf{V}$ into eight disjoint (and potentially empty) subsets. As shown in \citet{maasch2024local}, these eight partitions are defined by the types of paths that they share with $X$ and $Y$ (Table \ref{tab:partitions}). This basic property of cause-effect pairs in DAGs can be exploited for the design of local discovery algorithms, as previously used for adjustment set discovery \citep{maasch2024local} and causal fairness analysis \citep{maasch2025local}.  See Figures \ref{fig:partition_examples} and \ref{fig:z2} for visualizations of causal partitions on a full time DAG.

\begin{theorem}[Causal partitions (Theorem 3.1 in \citealt{maasch2024local})]  \label{theorem:partitions}
Let $\{X,Y\}$ be a cause-effect pair of interest in causal DAG $\g$. Let $Z$ be a third variable in $\g$. The eight partitions defined in Table~\ref{tab:partitions} are exhaustive and mutually exclusive, such that any $Z$ falls uniquely in one partition.
\end{theorem}

\paragraph{Partition Notation} We denote the union of two partitions in the form $\z_{5,7} \coloneqq \z_5 \cup \z_7$. When referring to a partition wrt a specific cause-effect pair $\{X,Y\}$, we use the notation $\z_1(X,Y)$. Partition labels inferred from data use hat notation (e.g., ground truth $\z_{1}$, inferred label $\widehat{\z}_{1}$). 




\section{Local-to-Global Time Series Discovery}
\label{sec:method}

This section introduces two algorithms: (1) IPaDL, a sound and complete yet naive extension of an existing parent discovery algorithm \citep{maasch2025local}, which served as an inspiration point and cheap baseline for this work; and (2) ITPD, a sound and complete constraint-based procedure that overcomes some of the inefficiencies of IPaDL. While IPaDL is a pure causal partitioning algorithm \citep{maasch2024local}, ITPD takes a combined strategy that leverages temporal structural priors and the local Markov condition (Definition \ref{def:local_markov}). To handle causal nonstationarity, ITPD does not assume a recurring window graph as done in baselines PCMCI \citep{runge2019detecting} and DYNOTEARS \citep{pamfil2020dynotears}, and instead learns the unique parent set for every node individually. 



\subsection{Data Settings: Stationary \& Nonstationary Data}
\label{sec:data_structure}

The proposed methods are applicable to two settings. While the nonstationary setting is the primary focus of this work, a stationary setting can be assumed if desired. Let $\mathbf{D}$ denote our observational dataset and $N$ the number of variables observed at each of $T$ time steps.
\begin{enumerate}
[label=\textbf{S\arabic*},leftmargin=\widthof{S1}+\labelsep,ref=S\arabic*] 
    \item \label{setting:single} \textit{Single multivariate time series with causal stationarity}: $\mathbf{D} \in \mathbb{R}^{T \times N}$. As in PCMCI, ITPD for \ref{setting:single} assumes causal stationarity over a maximum lag of $\tau_{\max}$ and returns a window graph.
    \item \label{setting:multi} \textit{Multiple multivariate time series with causal nonstationarity}: $\mathbf{D} \in \mathbb{R}^{M \times T \times N}$, where $M$ is the number of replicate time series. In \ref{setting:multi}, ITPD does not assume causal stationarity and returns a full time graph. Each $T \times N$ time series $\mathbf{T}$ is independently sampled $M$ times from the joint distribution: $\mathbf{T}_m \overset{\text{iid}}{\sim} p(\mathbf{V})$ for all $\{\mathbf{T}_m\}_{m=1}^M$. That is, each of these $M$ realizations represents the same underlying dynamical system. Under faithfulness, this implies that each trajectory is also an iid sample from the same SCM $\mathcal{M}$ with causal graph $\mathcal{G}$. 
\end{enumerate}
\ref{setting:single} can arise in finance and prediction markets (e.g., only a single ground truth trajectory can be observed for a given stock price). \ref{setting:multi} naturally arises in panel data sampled from a population, as commonly encountered in economics, sociology, epidemiology, and clinical research. In \ref{setting:multi}, replicate trajectories can represent patients in a clinical trial, citizens in census data, users of a wearable device, etc. Under \ref{setting:multi}, this work allows for causal nonstationarity. Natural examples where causal nonstationarity can arise include gene regulatory networks, where regulatory relationships change over development (e.g., in early embryonic stages versus senescence); ecological data, where populations can collapse and exotic species can be introduced; and climate data, where new biotic and abiotic causes can arise over geologic history (e.g., anthropogenic greenhouse gas emissions).

\begin{algorithm*}[!ht] 
\caption{Iterative Temporal Parent Discovery (ITPD).
} \label{alg:itpd}
    \begin{algorithmic}[1]
        \INPUT Time series $\mathbf{V} \in \mathbb{R}^{M \times T \times N}$ with $M$ observations, CI test \cit, significance level $\alpha$. Levels for a target at time step $t + 1$: $\alpha_1 = \max(\alpha, \min(\gamma, \kappa / (N t - 1)))$ for Phase I (Lines 8, 9, and 10), and $\alpha_2 = \alpha / (N t - 1)$ for Phase II (Line 14).

        \OUTPUT Adjacency list $\mathbf{A}$ where $\mathbf{A}[V^n_t] = \widehat{\pa}(V^n_t)$ for all $V^n_t \in \mathbf{V}$.

        \ASSUMPTIONS See \S\ref{sec:assumptions}. 
        
        \NOTATION \cit($\cdot$) denote Boolean queries. 

        \vspace{3mm}
        \STATE $\mathbf{A} \gets \{V_1^i \, \mapsto \, \varnothing : i \in [1,\dots,N]\}$ \COMMENT{Initialize roots: parent set is empty.}
        \FOR{$t \in [1, \dots, T-1]$}
            \FOR{$n \in [1, \dots, N]$}
                \STATE $X \gets V^n_{t}, Y \gets V^n_{t+1}, \z \gets \mathbf{V}_{1:t} \setminus \{X,Y\}$

                \medskip

                \textit{\color{Gray}Phase I: Identify a superset of $\pa(Y)$ in one to three CI tests per candidate. Sets $\widehat{\an}(\cdot)$ are obtained by  \ref{assumption:autocorrelation} or querying $\mathbf{A}$. Sets $\widehat{\pa}(\cdot)$ are obtained by querying $\mathbf{A}$. Level $\alpha_1$.}
                \STATE $\mathbf{S} \gets \{X\}$ \COMMENT{Initialize candidate set with known parent.}
                \FOR{$Z \in \z$ in descending time order}
                    \STATE \textbf{if} $Z \not\in  \widehat{\an}(\mathbf{S})$ \textbf{then} \COMMENT{By definition of $\z_8$.} \\
                        \STATE \quad \textbf{if} \cit($Z \ind Y$) \textbf{then} continue  \COMMENT{$Z \in \z_8 \not\subset \pa(Y)$.} \\
                    \STATE \textbf{if} \cit($Z \ind Y \mid \mathbf{S}$) \textbf{then} continue \COMMENT{$\mathbf{S}$ as it is when $Z$ is visited.}
                    \STATE \textbf{if} \cit($Z \not\ind Y \mid \widehat{\pa}(Z)$) \textbf{then} $\mathbf{S} \gets \mathbf{S} \cup \{Z\}$
                    \COMMENT{Local Markov (Def. \ref{def:local_markov}).}
                \ENDFOR

                \medskip

                \textit{\color{Gray}Phase II: Identify $\pa(Y)$ in one CI test per candidate. Level $\alpha_2$.}
                \STATE $\widehat{\pa}(Y) \gets \{X\}$ \COMMENT{Only the known parent, untested.}
                \STATE $\mathbf{S}' \gets \mathbf{S}$,
                $\mathbf{Z}' \gets \mathbf{S} \setminus \widehat{\pa}(Y)$
                \FOR{$Z \in \mathbf{Z}'$ in ascending time order}
                    \STATE \textbf{if} \cit($Z \not\ind Y \mid \mathbf{S}' \setminus \{Z\}$) \textbf{then}  $\widehat{\pa}(Y) \gets \widehat{\pa}(Y) \cup \{Z\}$
                    \STATE \textbf{else} $\mathbf{S}' \gets \mathbf{S}' \setminus \{Z\}$ \COMMENT{Remove non-parents.}
                \ENDFOR
                \STATE $\mathbf{A}[Y] \gets \widehat{\pa}(Y)$ \COMMENT{Update adjacency list.}
            \ENDFOR
        \ENDFOR
        \RETURN $\mathbf{A}$
    \end{algorithmic}
\end{algorithm*}

\subsection{A Naive Causal Partitioning Baseline}
\label{sec:ipadl_in_text}

This work was inspired by the premise that one could iteratively run an efficient local parent discovery procedure to obtain a global nonstationary graph. To this end, we devised a naive baseline method against which to compare our final algorithm: \underline{I}terative \underline{P}arent \underline{D}iscovery on \underline{L}eaves (IPaDL; Algorithm \ref{alg:itpd_naive}). Based on a subroutine of the local causal partitioning algorithm LD3 \citep{maasch2024local} that we term \underline{P}arent \underline{D}iscovery on \underline{L}eaves (PaDL, pronounced \textit{paddle}; Algorithm \ref{alg:padl}), IPaDL extends to the TiMINo setting for \ref{setting:single} and \ref{setting:multi}. Though sound and complete under sufficient conditions (Theorem \ref{theorem:naive_correct}), PaDL was designed for a setting with different structural priors: namely, no time-indexing and unknown relation between $X$ and $Y$. Thus, naively applying PaDL to each variable results in extraneous CI tests and unnecessarily large conditioning sets. IPaDL therefore serves as a proof-of-concept that iterative local parent discovery is a viable strategy for temporal graph learning, but thoughtful extensions are required for efficiency. See Appendix \ref{appendix:ipadl} for a complete discussion with proofs. To our knowledge, this work represents the first case of leveraging local causal partitioning for time series discovery. Additionally, this work uses causal partitioning as a convenient proof structure for proving soundness by cases.

\subsection{Iterative Temporal Parent Discovery}
\label{sec:itpd}

Incorporating lessons from the inefficiencies of IPaDL, we propose ITPD. While IPaDL is based purely on causal partitioning \citep{maasch2024local}, ITPD adopts a combined strategy to reduce reliance on higher-order CI tests. ITPD exploits (1) structural priors concerning temporal order and autocorrelation, (2) causal partitioning, and (3) the local Markov condition (Definition \ref{def:local_markov}). 

\subsubsection{Assumptions} 
\label{sec:assumptions}

ITPD assumes a TiMINo that satisfies the following conditions: acyclicity, jointly independent exogenous noise terms, local and global Markov, faithfulness, autocorrelation, and no contemporaneous links. We assume that variables in $\mathbf{V}_{2:T}$ are not latently confounded. Causal stationarity is not required, though a window graph of maximum lag $\tau_{\max}$ can be assumed if desired. We make no parametric assumptions, though the user-selected CI test may. We assume that all variables are measured at the same time steps $[1, \dots, T]$. We clarify the following graphical assumptions:

\begin{enumerate}[label=\textbf{A\arabic*}, ref=A\arabic*]
    \item \label{assumption:autocorrelation} \textit{Lag-1 autocorrelation assumption.} For all $\{V^n_{t}, V^n_{t+1}\}$, 
    $V^n_{t} \to V^n_{t+1}$.
    \item \label{assumption:no_contem} \textit{No contemporaneous links assumption.} For all $\{V^i_{t}, V^j_{t}\}$ where $i \neq j$, 
    $V^i_{t} \not\to V^j_{t}$. 
    \item \label{assumption:latent_variables} \textit{No latent confounding after $t=1$.} Variables in $\mathbf{V}_{2:T}$ are not latently confounded, though variables in $\mathbf{V}_{1}$ can be. This assumption accommodates the setting of \textit{unobserved history}: while $\mathbf{V}$ is measured from time $t=1$ forward, earlier time steps were not observed. Latent confounding is highly likely in the unobserved history, such that $\mathbf{V}_1$ are only roots in the observed graph.
\end{enumerate}
Table~\ref{tab:violations} shows performance impacts when \ref{assumption:no_contem} is violated.


\subsubsection{Method Description}

\paragraph{Data Settings} The experimental results and main contributions of this work revolve around setting \ref{setting:multi}. The main difference between \ref{setting:single} and \ref{setting:multi} is the data pre-processing performed prior to running ITPD. Namely, data corresponding to a single trajectory assumed to be stationary with maximum lag $\tau_{\max}$ can be sliced and stacked to obtain $M$ trajectories of length $\tau_{\max} +1$. Thus, with proper pre-processing, Algorithm \ref{alg:itpd} can be run as-is for either \ref{setting:single} or \ref{setting:multi}.

\paragraph{Significance Levels} 
ITPD performs CI tests in two phases. In Phase I (Lines 8–10), variables are screened to enter the candidate set $\mathbf{S}$, while Phase II (Line 14) determines which candidates are true parents. Since a removal in Phase I is irreversible, we use a more permissive threshold in Phase I and a stricter threshold in Phase II.
For a target variable $Y$ at time $t+1$, let $m=Nt-1$ denote the number of preceding variables considered as potential parents, excluding $X$. In Phase I, we set $\alpha_1 = \max\left(\alpha,\min\left(\gamma,\frac{\kappa}{m}\right)\right),$ where $\gamma>\alpha$ allows more permissive screening when $m$ is small, and $\kappa$ sets a budget for expected non-parents allowed into $\mathbf{S}$~\citep{gordon2007control}. As $m$ increases, the threshold decreases, with $\alpha$ serving as a lower bound to avoid overly conservative screening. In Phase II, we use $\alpha_2 = \frac{\alpha}{m},$
applying the Bonferroni correction~\citep{dunn1961multiple} to impose a stricter criterion for final parent selection. Details are deferred to Appendix~\ref{appendix:experimental_design}.

\paragraph{Sound \& Complete Iterative Parent Discovery} For each variable in $\mathbf{V}$, ITPD performs a biphasic procedure to identify its parents. To discover a superset of $\pa(Y)$, Phase I applies causal partitioning theory (Lines 7 and 8) and the local Markov condition (Line 10) using information learned at prior iterations. Phase II reduces this superset to strictly return the parent set. 

Let $X$ denote $V^n_t$ and $Y$ denote $V^n_{t+1}$, where we aim to discover the parents of $Y$. We initialize the candidate set $\mathbf{S}$ to be known parent $X$ per \ref{assumption:autocorrelation} (Line 5). We evaluate all $Z \in \mathbf{V}_{1:t}$ in descending time order, which demonstrated reduced cardinality of $\mathbf{S}$ and significantly fewer CI tests in our ablation study (Table \ref{tab:ablation}). ITPD winnows out $\z_8(X,Y)$ with a marginal independence test, as this set is disjoint with $\pa(Y)$ (Line 8). To avoid unnecessary tests, Line 8 is gated by an ancestry test that queries the growing adjacency list.

Next, ITPD tests whether $Z$ is independent of $Y$ given the growing $\mathbf{S}$ (Line 9): if $Z$ is independent of $Y$ given a (likely incomplete) superset of its parents, then $Z$ is not a parent. Line 9 is the forward step of IAMB and Grow-Shrink \citep{margaritis1999bayesian}, run once per candidate in reverse temporal order and without a ranking by dependence strength. As $\mathbf{S}$ is likely incomplete, Line 10 performs a local Markov test to further reduce $|\mathbf{S}|$: if $Z$ is independent of $Y$ given its previously learned parent set, then $Y$ is not a descendant of $Z$ (Definition \ref{def:local_markov}). Phase II removes non-parents from the superset discovered by Phase I, yielding an exact algorithm for parent discovery. 

We show that Algorithm \ref{alg:itpd} is sound (i.e., outputs are correct) and complete (i.e., outputs are maximally informative) under an oracle and the assumptions given in \S\ref{sec:assumptions}. We first prove that the local discovery subroutine of ITPD is sound and complete for a single variable. We then prove by induction that ITPD is sound and complete for the global graph. Proofs are in Appendix \ref{appendix:correctness_itpd}.

\begin{theorem}[Sound and complete local parent discovery]
\label{theorem:sound_complete_local}
    Under sufficient conditions (\S\ref{sec:assumptions}, \S\ref{appendix:correctness_itpd}) and an independence oracle, ITPD is locally sound: Lines 5–16 of Algorithm \ref{alg:itpd} return exactly the correct parent set for a given variable $Y \in \mathbf{V}^n_{t>1}$. 
    Further, ITPD is locally complete: inferences returned by Lines 5–16 of Algorithm \ref{alg:itpd} are guaranteed to be at least as informative as those returned by any other parent discovery algorithm under these conditions. 
\end{theorem}

\begin{theorem}[Sound and complete global discovery]
\label{theorem:sound_complete}
    Under sufficient conditions (\S\ref{sec:assumptions}) and an independence oracle, ITPD is sound: Algorithm \ref{alg:itpd} returns the correct global time series DAG. Further, ITPD is complete: Algorithm \ref{alg:itpd} is guaranteed to be at least as informative as any other global causal discovery algorithm under these conditions.
\end{theorem}




\begin{theorem}[Polynomial computational complexity]
\label{theorem:complexity}
    Given a polynomial time CI test, ITPD executes in polynomial time with $O(N^2 T^2)$ CI tests under \ref{setting:multi} and  $O(N^2\tau_{\max}^2)$ under \ref{setting:single}. 
\end{theorem}

\subsubsection{Ablation Post-Mortem} 

To illustrate the value of the strategy presented in Algorithm \ref{alg:itpd}, we performed a systematic ablation (Table \ref{tab:ablation}). The results of step-wise ablations were used to justify our design choices on the final draft of Algorithm \ref{alg:itpd}. Along the way, several additional causal partitioning steps were introduced and phased out due to low yield. This included steps that tested for $\z_4$ and $\z_5$. While Algorithm \ref{alg:itpd} remained sound and complete under these additional partitioning steps, their hypothesized added value wrt reducing mean conditioning set size did not pan out in practice. Hence, the only steps that incorporate causal partitioning are Lines 7 and 8. See Appendix \ref{appendix:ablations} for further insights.

\subsubsection{Formal Verification}
\label{sec:autoform}

We used AI-assisted autoformalization to express select theoretical results as machine-checkable statements and proofs in Lean~4.33.1 \citep{moura2021lean}. Autoformalization was performed by Codex and Claude.\footnote{\href{https://chatgpt.com/codex/}{\texttt{https://chatgpt.com/codex}}, \href{https://claude.ai/}{\texttt{https://claude.ai/}}} See Appendix \ref{appendix:autoform} and GitHub\footnote{\href{https://anonymous.4open.science/r/itpd-0FC0/README.md}{\texttt{https://anonymous.4open.science/r/itpd-0FC0}}} for full details on our methods and results.


\begin{table}[t]
    \centering
    \footnotesize
    \setlength{\tabcolsep}{2.8pt}
    \begin{tabular}{l ccccc c ccccc c ccccc}
    \toprule
    & \multicolumn{5}{c}{Unique CI tests per candidate} && \multicolumn{5}{c}{Mean conditioning set} && \multicolumn{5}{c}{Max conditioning set} \\
    \cmidrule(lr){2-6} \cmidrule(lr){8-12} \cmidrule(lr){14-18}
    Method & $d = 1$ & 2 & 3 & 5 & 8 && $d = 1$ & 2 & 3 & 5 & 8 && $d = 1$ & 2 & 3 & 5 & 8 \\
    \midrule
    ITPD & \textbf{1.23} & \textbf{1.41} & \textbf{1.32} & \textbf{1.26} & \textbf{1.26} && \textbf{0.7} & \textbf{2.0} & \textbf{3.1} & \textbf{5.0} & \textbf{7.7} && \textbf{4} & \textbf{6} & \textbf{7} & \textbf{10} & \textbf{13} \\
    IAMB & 6.02 & 11.83 & 14.23 & 16.34 & 17.67 && 5.6 & 8.0 & 8.6 & 9.2 & 9.7 && 15 & \underline{20} & \underline{20} & \underline{20} & \underline{20} \\
    IPaDL & \underline{1.76} & \underline{2.74} & \underline{2.98} & \underline{3.05} & \underline{3.06} && 21.7 & 58.0 & 65.3 & 67.3 & 67.5 && 211 & 293 & 298 & 299 & 299 \\
    FullCI & 1.00 & 1.00 & 1.00 & 1.00 & 1.00 && 206 & 206 & 206 & 206 & 206 && 299 & 299 & 299 & 299 & 299 \\
    PCMCI & -- & -- & -- & -- & -- && \underline{4.2} & \underline{7.6} & \underline{8.3} & \underline{8.9} & \underline{9.5} && \underline{14} & \underline{20} & 22 & 23 & 27 \\
    \bottomrule
    \end{tabular}
    \caption{Unique CI tests per candidate and conditioning set sizes under the exact oracle on stationary graphs with $N = 20$, $T = 16$, and lag 1. All methods were tested on 200 graphs per column. Under the exact oracle, IAMB cannot rank candidates by dependence strength.}
    \label{tab:oracle}
\end{table}


\section{Empirical Results}
\label{sec:empirics}

\subsection{Experimental Design}

We empirically validated ITPD and IPaDL by comparing to baselines on the basis of: (1) F1 scores relative to the ground truth graph; (2) total CI tests used; (3) average order of CI tests; and (4) wall-clock time. We chose $\alpha=0.01, \gamma = 0.2, \kappa = 50$ in all experiments. Supplemental results and additional experimental design details are in Appendix \ref{appendix:results}, including notes on compute resources.  

\paragraph{Baselines} IAMB \citep{tsamardinos2003algorithms}, IPaDL  (Algorithm \ref{alg:itpd_naive}), and FullCI search the same candidates per target as ITPD. Our use of IAMB is analogous to IPaDL, in the sense that we iteratively apply a local learning procedure that is not specifically tailored to the current setting. Lasso \citep{tibshirani1996regression} fits one fixed-penalty regression per target. DF-PC \citep{kim2026deduce}, PCMCI \citep{runge2018causal}, DYNOTEARS \citep{pamfil2020dynotears}, and CD-NOD \citep{huang2020causal} were run using published code. PCMCI, CD-NOD, and DYNOTEARS pool all time steps into one window graph (maximum lag 1 unless stated). With Fisher-z on these data, CD-NOD acted as the time-ordered PC algorithm. Each baseline with a significance level was run at $\alpha$ and at $\alpha$ divided by its number of candidates, as in Line 14 of ITPD. We report the level with the higher mean F1. 


\paragraph{Data Generation \& Evaluation} In the \textit{main setting}, data are $M$ linear-Gaussian observations of $N = 20$ variables over $T = 16$ time steps from nonstationary graphs with lag-1 edges and in-degree $d$. CI tests are Fisher-z ($\alpha = 0.01$). Results are means over 20 graphs. F1 is computed on common targets (i.e., those that every method can test). Unique CI tests per candidate is the number of distinct CI tests executed in a given run, divided by the total number of candidates. A method is higher or lower than ITPD when the 95\% bootstrap interval of the difference paired by graph (10{,}000 resamples) excludes zero, and tied otherwise. In tables, bold marks the best and underline the second best value of a column. A dash marks a method with no comparable count of CI tests per candidate. FullCI, which always tests each candidate once given all others, is unranked on tests.

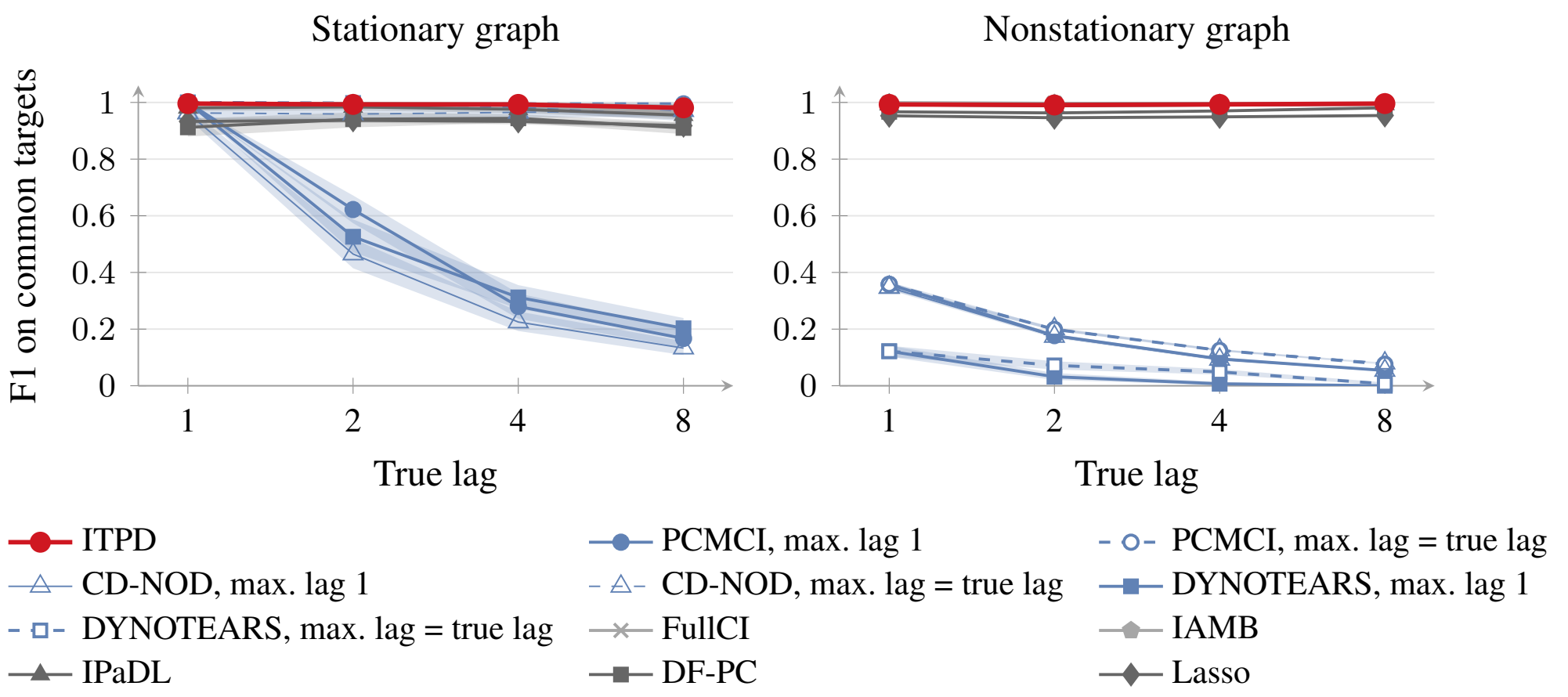
\begin{figure}[t] 
    \centering
\begin{tikzpicture}[scale=0.9]
\begin{axis}[
    name=lagA, 
    resaxis, width=0.47\linewidth, height=4.4cm,
    xmin=-0.3, xmax=3.3, ymin=0, ymax=1.06,
    xtick={0,1,2,3}, xticklabels={1,2,4,8}, ytick={0,0.2,0.4,0.6,0.8,1},
    xlabel={True lag}, ylabel={F1 on common targets},
    title={Stationary graph}, legend columns=3, legend style={at={(1.08,-0.42)}, anchor=north},
]
\addlegendimage{mITPD} \addlegendentry{ITPD}
\addplot[fill=glaucous, fill opacity=0.22, draw=none, no markers, forget plot] coordinates {(0,1) (1,0.67176) (2,0.32593) (3,0.19911) (3,0.13507) (2,0.23622) (1,0.57378) (0,1)} -- cycle;
\addplot[fill=glaucous, fill opacity=0.22, draw=none, no markers, forget plot] coordinates {(0,1) (1,1) (2,1) (3,1) (3,0.98976) (2,0.98415) (1,0.98393) (0,1)} -- cycle;
\addplot[fill=glaucous, fill opacity=0.22, draw=none, no markers, forget plot] coordinates {(0,0.97958) (1,0.51915) (2,0.26242) (3,0.15728) (3,0.10833) (2,0.1933) (1,0.4147) (0,0.94419)} -- cycle;
\addplot[fill=glaucous, fill opacity=0.22, draw=none, no markers, forget plot] coordinates {(0,0.97958) (1,0.98358) (2,0.98205) (3,0.98817) (3,0.95016) (2,0.94599) (1,0.92962) (0,0.94419)} -- cycle;
\addplot[fill=glaucous, fill opacity=0.22, draw=none, no markers, forget plot] coordinates {(0,1) (1,0.5874) (2,0.35533) (3,0.23916) (3,0.16566) (2,0.27192) (1,0.47091) (0,1)} -- cycle;
\addplot[fill=glaucous, fill opacity=0.22, draw=none, no markers, forget plot] coordinates {(0,1) (1,1) (2,0.99635) (3,0.97505) (3,0.93428) (2,0.97009) (1,0.99386) (0,1)} -- cycle;
\addplot[fill=black!55, fill opacity=0.22, draw=none, no markers, forget plot] coordinates {(0,0.99814) (1,0.99774) (2,0.99572) (3,0.97723) (3,0.94488) (2,0.98283) (1,0.99465) (0,0.99661)} -- cycle;
\addplot[fill=black!55, fill opacity=0.22, draw=none, no markers, forget plot] coordinates {(0,0.99838) (1,0.99806) (2,0.99787) (3,0.99464) (3,0.97581) (2,0.99073) (1,0.99593) (0,0.99628)} -- cycle;
\addplot[fill=black!55, fill opacity=0.22, draw=none, no markers, forget plot] coordinates {(0,0.99122) (1,0.99388) (2,0.98774) (3,0.96913) (3,0.93949) (2,0.96257) (1,0.97201) (0,0.96978)} -- cycle;
\addplot[fill=black!55, fill opacity=0.22, draw=none, no markers, forget plot] coordinates {(0,0.94239) (1,0.96476) (2,0.95836) (3,0.92974) (3,0.88826) (2,0.92849) (1,0.91201) (0,0.87956)} -- cycle;
\addplot[fill=black!55, fill opacity=0.22, draw=none, no markers, forget plot] coordinates {(0,0.94508) (1,0.94549) (2,0.94273) (3,0.92862) (3,0.90766) (2,0.92257) (1,0.92855) (0,0.92094)} -- cycle;
\addplot[fill=parents, fill opacity=0.22, draw=none, no markers, forget plot] coordinates {(0,0.99869) (1,0.99762) (2,0.99726) (3,0.99025) (3,0.96966) (2,0.98677) (1,0.98489) (0,0.99218)} -- cycle;
\addplot[mPCMCI] coordinates {(0,1.000) (1,0.622) (2,0.280) (3,0.167)}; \addlegendentry{PCMCI, max.\ lag 1}
\addplot[mPCMCI, dashed, mark options={solid, fill=white}] coordinates {(0,1.000) (1,0.993) (2,0.995) (3,0.995)}; \addlegendentry{PCMCI, max.\ lag = true lag}
\addplot[glaucous, thin, mark=triangle, mark size=3.0pt, mark options={solid}] coordinates {(0,0.963) (1,0.465) (2,0.225) (3,0.133)}; \addlegendentry{CD-NOD, max.\ lag 1}
\addplot[glaucous, thin, mark=triangle, mark size=3.0pt, dashed, mark options={solid, fill=white}] coordinates {(0,0.963) (1,0.959) (2,0.965) (3,0.971)}; \addlegendentry{CD-NOD, max.\ lag = true lag}
\addplot[mDYNO] coordinates {(0,1.000) (1,0.526) (2,0.312) (3,0.203)}; \addlegendentry{DYNOTEARS, max.\ lag 1}
\addplot[mDYNO, dashed, mark options={solid, fill=white}] coordinates {(0,1.000) (1,0.998) (2,0.986) (3,0.956)}; \addlegendentry{DYNOTEARS, max.\ lag = true lag}
\addplot[mFullCI] coordinates {(0,0.997) (1,0.996) (2,0.991) (3,0.962)}; \addlegendentry{FullCI}
\addplot[mIAMB] coordinates {(0,0.997) (1,0.997) (2,0.995) (3,0.986)}; \addlegendentry{IAMB}
\addplot[mIPaDL] coordinates {(0,0.981) (1,0.984) (2,0.976) (3,0.955)}; \addlegendentry{IPaDL}
\addplot[mDFPC] coordinates {(0,0.912) (1,0.941) (2,0.944) (3,0.910)}; \addlegendentry{DF-PC}
\addplot[mLasso] coordinates {(0,0.933) (1,0.937) (2,0.933) (3,0.918)}; \addlegendentry{Lasso}
\addplot[mITPD] coordinates {(0,0.996) (1,0.993) (2,0.993) (3,0.981)};
\end{axis}
\begin{axis}[
     at={(lagA.south east)}, anchor=south west, xshift=1.0cm,
    resaxis, width=0.47\linewidth, height=4.4cm,
    xmin=-0.3, xmax=3.3, ymin=0, ymax=1.06,
    xtick={0,1,2,3}, xticklabels={1,2,4,8}, ytick={0,0.2,0.4,0.6,0.8,1},
    xlabel={True lag}, 
    title={Nonstationary graph}, 
]
\addplot[fill=glaucous, fill opacity=0.22, draw=none, no markers, forget plot] coordinates {(0,0.36643) (1,0.18225) (2,0.099784) (3,0.057158) (3,0.050472) (2,0.090278) (1,0.17214) (0,0.35028)} -- cycle;
\addplot[fill=glaucous, fill opacity=0.22, draw=none, no markers, forget plot] coordinates {(0,0.36643) (1,0.20355) (2,0.12863) (3,0.077335) (3,0.074561) (2,0.12133) (1,0.19448) (0,0.35028)} -- cycle;
\addplot[fill=glaucous, fill opacity=0.22, draw=none, no markers, forget plot] coordinates {(0,0.35558) (1,0.18021) (2,0.098276) (3,0.057186) (3,0.050487) (2,0.087735) (1,0.16826) (0,0.33816)} -- cycle;
\addplot[fill=glaucous, fill opacity=0.22, draw=none, no markers, forget plot] coordinates {(0,0.35558) (1,0.20594) (2,0.12947) (3,0.081362) (3,0.078221) (2,0.12246) (1,0.19525) (0,0.33816)} -- cycle;
\addplot[fill=glaucous, fill opacity=0.22, draw=none, no markers, forget plot] coordinates {(0,0.14147) (1,0.045188) (2,0.01352) (3,0) (3,0) (2,0.0021845) (1,0.020358) (0,0.10213)} -- cycle;
\addplot[fill=glaucous, fill opacity=0.22, draw=none, no markers, forget plot] coordinates {(0,0.14147) (1,0.086551) (2,0.058949) (3,0.011743) (3,0.0022143) (2,0.039302) (1,0.057362) (0,0.10213)} -- cycle;
\addplot[fill=black!55, fill opacity=0.22, draw=none, no markers, forget plot] coordinates {(0,0.99902) (1,0.99832) (2,0.99722) (3,0.99763) (3,0.99584) (2,0.99502) (1,0.9958) (0,0.99778)} -- cycle;
\addplot[fill=black!55, fill opacity=0.22, draw=none, no markers, forget plot] coordinates {(0,0.99934) (1,0.9977) (2,0.99823) (3,0.99815) (3,0.99599) (2,0.9963) (1,0.9961) (0,0.99814)} -- cycle;
\addplot[fill=black!55, fill opacity=0.22, draw=none, no markers, forget plot] coordinates {(0,0.99398) (1,0.99138) (2,0.99356) (3,0.99403) (3,0.99101) (2,0.98795) (1,0.98408) (0,0.98875)} -- cycle;
\addplot[fill=black!55, fill opacity=0.22, draw=none, no markers, forget plot] coordinates {(0,0.97019) (1,0.96655) (2,0.97324) (3,0.98246) (3,0.97899) (2,0.96697) (1,0.95846) (0,0.96312)} -- cycle;
\addplot[fill=black!55, fill opacity=0.22, draw=none, no markers, forget plot] coordinates {(0,0.95601) (1,0.94909) (2,0.9532) (3,0.95756) (3,0.94989) (2,0.9449) (1,0.94205) (0,0.95016)} -- cycle;
\addplot[fill=parents, fill opacity=0.22, draw=none, no markers, forget plot] coordinates {(0,0.99561) (1,0.99209) (2,0.99533) (3,0.99702) (3,0.99517) (2,0.99119) (1,0.98952) (0,0.99112)} -- cycle;
\addplot[mPCMCI] coordinates {(0,0.358) (1,0.177) (2,0.095) (3,0.054)};
\addplot[mPCMCI, dashed, mark options={solid, fill=white}] coordinates {(0,0.358) (1,0.199) (2,0.125) (3,0.076)};
\addplot[glaucous, thin, mark=triangle, mark size=3.0pt, mark options={solid}] coordinates {(0,0.347) (1,0.174) (2,0.093) (3,0.054)};
\addplot[glaucous, thin, mark=triangle, mark size=3.0pt, dashed, mark options={solid, fill=white}] coordinates {(0,0.347) (1,0.201) (2,0.126) (3,0.080)};
\addplot[mDYNO] coordinates {(0,0.122) (1,0.032) (2,0.007) (3,0.000)};
\addplot[mDYNO, dashed, mark options={solid, fill=white}] coordinates {(0,0.122) (1,0.072) (2,0.049) (3,0.007)};
\addplot[mFullCI] coordinates {(0,0.998) (1,0.997) (2,0.996) (3,0.997)};
\addplot[mIAMB] coordinates {(0,0.999) (1,0.997) (2,0.997) (3,0.997)};
\addplot[mIPaDL] coordinates {(0,0.992) (1,0.988) (2,0.991) (3,0.993)};
\addplot[mDFPC] coordinates {(0,0.967) (1,0.963) (2,0.970) (3,0.981)};
\addplot[mLasso] coordinates {(0,0.953) (1,0.946) (2,0.949) (3,0.954)};
\addplot[mITPD] coordinates {(0,0.993) (1,0.991) (2,0.993) (3,0.996)};
\end{axis}
\end{tikzpicture}
    \caption{Mean F1 on common targets against the true lag on stationary graphs (left) and nonstationary graphs (right) (linear-Gaussian data, $N = 10$, $T = 16$, and $T = 24$ at true lag 8; $d = 2$, $M = 2000$; 20 graphs per lag). 
    PCMCI and DYNOTEARS ran with maximum lag 1 (solid) and with the maximum lag set to the true lag (dashed, hollow marks). Each point is a mean over 20 graphs, and the shaded bands are 95\% intervals. 
    }
    \label{fig:lag}
\end{figure}

\subsection{Results \& Discussion}

\paragraph{Oracle Validation} ITPD and IPaDL returned the true graph in all 32 exact-oracle settings (4240 graphs in total) with $N$ from 20 to 80, $T$ from 16 to 50, and a lag of 1 or 2 (Table~\ref{tab:oracle}). In every setting, ITPD ran fewer unique CI tests per candidate than IAMB and IPaDL and had smaller mean and maximum conditioning sets than all methods. IPaDL had significantly improved mean conditioning set size relative to FullCI. However, maximum conditioning set size was $22.9\times$ to $52.2\times$ greater for IPaDL than ITPD, highlighting the success of our core method over our naive baseline.

\paragraph{Synthetic Data Results} In the main setting, ITPD outperformed Lasso, PCMCI, CD-NOD, and DYNOTEARS in every setting wrt mean F1 score (Table~\ref{tab:all}). FullCI outperformed ITPD at $d \in \{3, 5\}$ and $M = 2000$ (by up to 0.007). Against IAMB, ITPD outperformed in 2 of 12 settings, tied in 4, and underperformed in 6 ($-0.042$ to $+0.004$). With nonlinear causal mechanisms and a nonlinear CI test (generalized covariance measure with boosted regressions; \citealt{shah2020hardness}), ITPD was first or second most performant in 6 of 8 settings (Table~\ref{tab:nonlinear}). IAMB was first or second in 3 of 8 settings, IPaDL in two, and PCMCI in one. DYNOTEARS was first or second in stationary graphs only.  
At $M = 500$, ITPD sat on the Pareto frontier wrt F1 and total CI tests (Figure \ref{fig:frontier_tests}). Across 76 finite data settings, ITPD ran 0.10$\times$ to 0.86$\times$ the number of unique CI tests per candidate as IAMB (Tables~\ref{tab:tests}, \ref{tab:scale}). At $M = 2000$, IPaDL and DF-PC ran fewer tests than ITPD only at $d = 1$. 
Counter to our oracle results, IAMB had smaller conditioning sets. 

\paragraph{Robustness to Nonstationarity} As ITPD does not assume causal stationarity, F1 scores remain robust under both stationary and nonstationary conditions (Figure \ref{fig:lag}). In the stationary setting, PCMCI and DYNOTEARS start strong but experience a steep performance decay as the true maximum lag departs from the assumed lag. In the nonstationary setting, these algorithms start low and end low. In nonstationary graphs with 5\% of edges randomly redrawn at each time step, F1 scores for PCMCI, DYNOTEARS, and CD-NOD declined to 0.562 to 0.808, while ITPD consistently maintained 0.984 to 0.994 F1 (Figure~\ref{fig:changing}). IAMB outperformed ITPD in all 16 settings by a margin of by 0.003 to 0.064 ($d \in \{2, 5\}$, $M \in \{500, 2000\}$). These observations emphasize the brittleness of algorithms that require unverifiable stationarity assumptions.

\begin{figure}[t] 
    \centering

\begin{tikzpicture}[scale=0.9]
\begin{axis}[
    name=frontA, 
    resaxis, width=0.52\linewidth, height=5.2cm,
    xmode=log, xmin=0.3, xmax=150, ymin=0.7, ymax=1.02,
    xtick={0.5,1,2,5,10,20,50,100}, log ticks with fixed point, ytick={0.7,0.8,0.9,1},
    xlabel={Seconds per run (log scale)}, ylabel={F1 on common targets},
    title={$d = 2$}, clip=false
]
\addplot[black!55, thick, dashed] coordinates {(0.63,0.963) (1.27,0.963) (1.27,0.967) (3.66,0.967) (3.66,0.992) (6.18,0.992) (6.18,0.994) (150,0.994)};
\addplot[only marks, mark=*, mark size=3pt, parents] coordinates {(3.66,0.992)};
\node[anchor=north, inner sep=4pt, font=\footnotesize\bfseries, text=parents] at (axis cs:3.66,0.992) {ITPD};
\addplot[only marks, mark=*, mark size=2.2pt, glaucous] coordinates {(6.18,0.994)};
\node[anchor=west, inner sep=4pt, font=\scriptsize] at (axis cs:6.18,0.994) {IAMB};
\addplot[only marks, mark=*, mark size=2.2pt, glaucous] coordinates {(1.27,0.967)};
\node[anchor=north, inner sep=4pt, font=\scriptsize] at (axis cs:1.27,0.967) {IPaDL};
\addplot[only marks, mark=*, mark size=2.2pt, glaucous] coordinates {(33.96,0.935)};
\node[anchor=north, inner sep=4pt, font=\scriptsize] at (axis cs:33.96,0.935) {FullCI};
\addplot[only marks, mark=o, thick, mark size=2.2pt, glaucous] coordinates {(9.91,0.926)};
\node[anchor=north, inner sep=4pt, font=\scriptsize] at (axis cs:9.91,0.926) {DF-PC};
\addplot[only marks, mark=triangle*, mark options={rotate=180}, mark size=2.6pt, glaucous] coordinates {(40.40,0.708)};
\node[anchor=south, inner sep=4pt, font=\scriptsize] at (axis cs:40.40,0.708) {PCMCI (0.205)};
\addplot[only marks, mark=diamond*, mark size=2.2pt, glaucous] coordinates {(0.63,0.963)};
\node[anchor=north, inner sep=4pt, font=\scriptsize] at (axis cs:0.63,0.963) {Lasso};
\addplot[only marks, mark=triangle*, mark options={rotate=180}, mark size=2.6pt, glaucous] coordinates {(1.57,0.708)};
\node[anchor=south, inner sep=4pt, font=\scriptsize] at (axis cs:1.57,0.708) {DYNOTEARS (0.045)};
\end{axis}
\begin{axis}[
     at={(frontA.south east)}, anchor=south west, xshift=1.2cm,
    resaxis, width=0.52\linewidth, height=5.2cm,
    xmode=log, xmin=0.3, xmax=150, ymin=0.7, ymax=1.02,
    xtick={0.5,1,2,5,10,20,50,100}, log ticks with fixed point, ytick={0.7,0.8,0.9,1},
    xlabel={Seconds per run (log scale)}, 
    title={$d = 5$}, clip=false
]
\addplot[black!55, thick, dashed] coordinates {(0.60,0.916) (4.01,0.916) (4.01,0.961) (12.07,0.961) (12.07,0.972) (150,0.972)};
\addplot[only marks, mark=*, mark size=3pt, parents] coordinates {(4.01,0.961)};
\node[anchor=north, inner sep=4pt, font=\footnotesize\bfseries, text=parents] at (axis cs:4.01,0.961) {ITPD};
\addplot[only marks, mark=*, mark size=2.2pt, glaucous] coordinates {(12.07,0.972)};
\node[anchor=west, inner sep=4pt, font=\scriptsize] at (axis cs:12.07,0.972) {IAMB};
\addplot[only marks, mark=*, mark size=2.2pt, glaucous] coordinates {(2.31,0.913)};
\node[anchor=north, inner sep=4pt, font=\scriptsize] at (axis cs:2.31,0.913) {IPaDL};
\addplot[only marks, mark=*, mark size=2.2pt, glaucous] coordinates {(33.01,0.815)};
\node[anchor=north, inner sep=4pt, font=\scriptsize] at (axis cs:33.01,0.815) {FullCI};
\addplot[only marks, mark=o, thick, mark size=2.2pt, glaucous] coordinates {(28.29,0.837)};
\node[anchor=south, inner sep=4pt, font=\scriptsize] at (axis cs:28.29,0.837) {DF-PC};
\addplot[only marks, mark=triangle*, mark options={rotate=180}, mark size=2.6pt, glaucous] coordinates {(53.58,0.708)};
\node[anchor=south, inner sep=4pt, font=\scriptsize] at (axis cs:53.58,0.708) {PCMCI (0.364)};
\addplot[only marks, mark=diamond*, mark size=2.2pt, glaucous] coordinates {(0.60,0.916)};
\node[anchor=north, inner sep=4pt, font=\scriptsize] at (axis cs:0.60,0.916) {Lasso};
\addplot[only marks, mark=triangle*, mark options={rotate=180}, mark size=2.6pt, glaucous] coordinates {(1.71,0.708)};
\node[anchor=south, inner sep=4pt, font=\scriptsize] at (axis cs:1.71,0.708) {DYNOTEARS (0.062)};
\end{axis}
\end{tikzpicture}
    
    \caption{ITPD lies on the Pareto frontier (dashed line) wrt mean F1 and wall-clock time. $M = 500$ and $d \in \{2,5\}$, T=16, 200 graphs per methods. Triangles are methods with F1 below the axis, given in the label. The dashed line is the Pareto frontier. Each point is a mean over 20 graphs. 
    }
    \label{fig:frontier}
\end{figure}
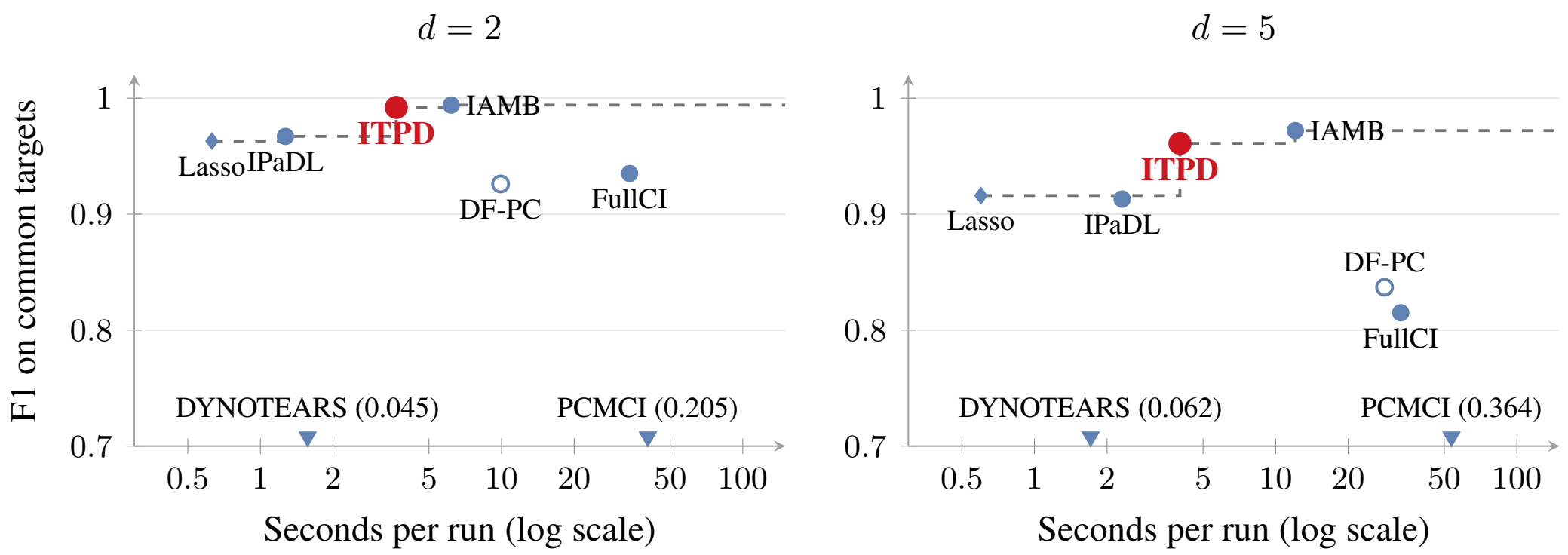



\begin{figure}[t]
    \centering
    \includegraphics[width=0.8\linewidth]{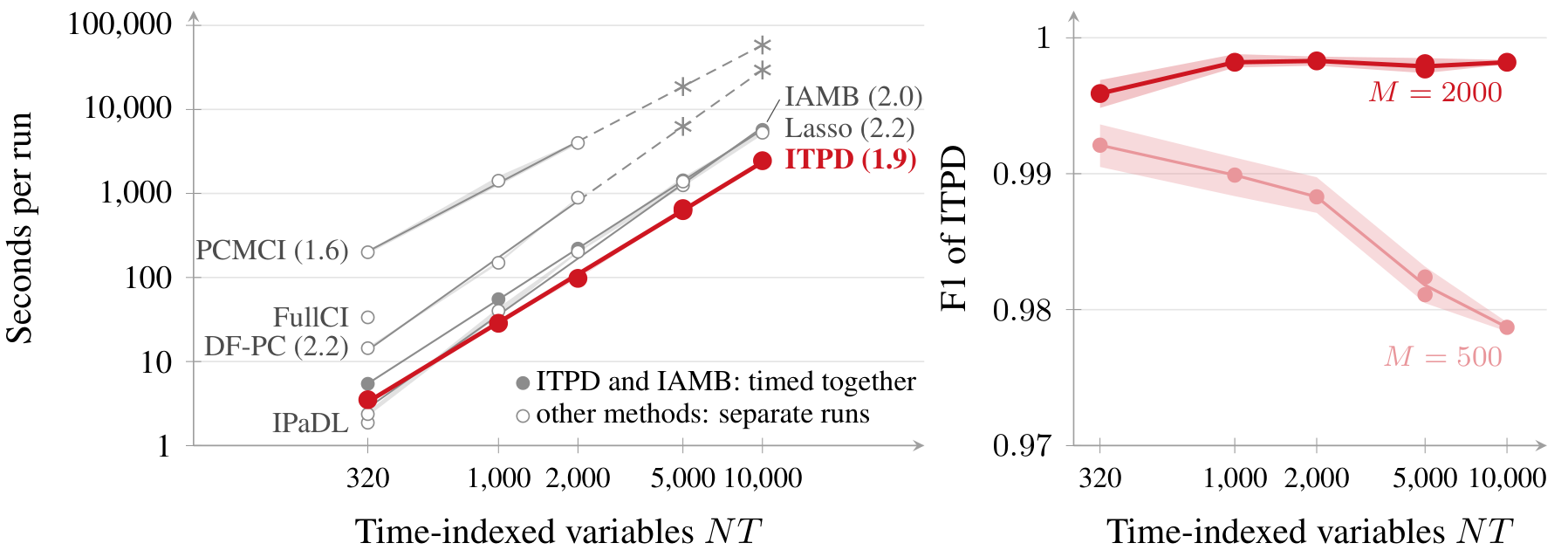}
    \caption{Seconds per run (left) and F1 of ITPD (right) versus total time-indexed variables, using Fisher-z CI tests at $M = 2000$ and $d = 2$. Left: For each method, numbers in parentheses give the fitted exponent for the power by which wall-clock time scaled with total variables. Open marks are methods timed in separate runs on shared nodes. Each point is a mean over 10 graphs, and the shaded bands are 95\% intervals. Asterisks are estimates from the fitted power law at sizes where the method was not run.}
    \label{fig:scaling}
\end{figure}

\paragraph{Scalability} ITPD lies on the Pareto frontier wrt F1 and wall-clock time in the main setting (Figure \ref{fig:frontier}). To assess scalability, we compared ITPD F1 and wall-clock time to baselines as graph size (Figure~\ref{fig:scaling}, Table \ref{tab:scale}) and sample size increased (Figure \ref{fig:samples}). From 1000 to 2000 variables, ITPD was the fastest running method. Total seconds per run grew with the number of time-indexed variables $NT$ as a power law with exponent 1.9 for ITPD and 2.0 for IAMB (Figure~\ref{fig:scaling}, left). As graph size scaled, ITPD maintained an F1 of 0.997 to 0.998 at $M = 2000$ (Figure~\ref{fig:scaling}, right). 
ITPD performs approximately as well or better than baselines as sample size scales in the main setting for $M \geq 100$ (Figure \ref{fig:samples}). Owing to large conditioning sets, FullCI begins weak but catches up to the most performant algorithms as sample size increases. PCMCI and DYNOTEARS do not display the same improvements, owing to insurmountable bias introduced by false stationarity assumptions. 


\paragraph{Gene Regulatory Networks} On DREAM4 gene regulatory networks 3 and 5 \citep{marbach2010revealing} re-simulated with GeneNetWeaver \citep{schaffter2011genenetweaver}, ITPD outperformed IAMB in all 6 settings (Table~\ref{tab:gnw}). Against the baseline with the highest mean F1 in each setting, ITPD scored lower in 4 settings, tied in 1, and outperformed in 1. PCMCI, CD-NOD, and DYNOTEARS outperformed ITPD in 6 of 6 comparisons at $M = 200$, where F1 covers the early time steps only.

\subsection{Limitations \& Future Directions} 

While ITPD occupies a favorable position on the Pareto frontier wrt F1 and wall-clock time and F1 and total CI tests, it does not uniformly outcompete alternative methods in all settings. A lack of public, real-world benchmark data for time series in the favorable regime for ITPD inhibited evaluation on real data. Developing such benchmarks would be a significant contribution to the temporal causal discovery community. Future work could improve on the real-world practicality of ITPD by permitting latent variables and contemporaneous links.

\bibstyle{plainnat}
\bibliography{references}

\clearpage



\clearpage
\appendix

\addcontentsline{toc}{section}{Supplementary Materials} 
\part{Supplementary Materials} 
\begin{spacing}{0.65} 
\parttoc 
\end{spacing}

\counterwithin{figure}{section} 
\counterwithin{table}{section}
\counterwithin{algorithm}{section}

\clearpage

\section{AI Use Statement} 

This work used generative AI for the following tasks: (1) refining a subset of our conceptual framework; (2) refining hypotheses; (3) generating synthetic data sets; (4) designing and implementing finite data experiments; (5) generating  software code or refactoring human-written code; (6) Lean autoformalization of manually derived theoretical results; and (7) error-checking manually derived proofs. The authors \textbf{did not} use generative AI for initial conceptualization of any algorithms (nor for any stage of developing IPaDL); writing or editing natural language text in the main body, though AI was used for text generation related to autoformalization (Appendix \ref{sec:autoform}); writing or editing pseudocode; or bibliography generation. Systems used included Claude Fable 5.1 and 5.2 (Anthropic), GPT-6 Astra (OpenAI), and Gemini 4 Carbon (Google).


\section{Proofs: ITPD (Algorithm \ref{alg:itpd})}


\subsection{Soundness and Completeness}
\label{appendix:correctness_itpd}

\definecolor{A_color}{HTML}{FDD5B1}
\definecolor{B_color}{HTML}{CCCCFF}
\definecolor{C_color}{HTML}{FDDDE6}

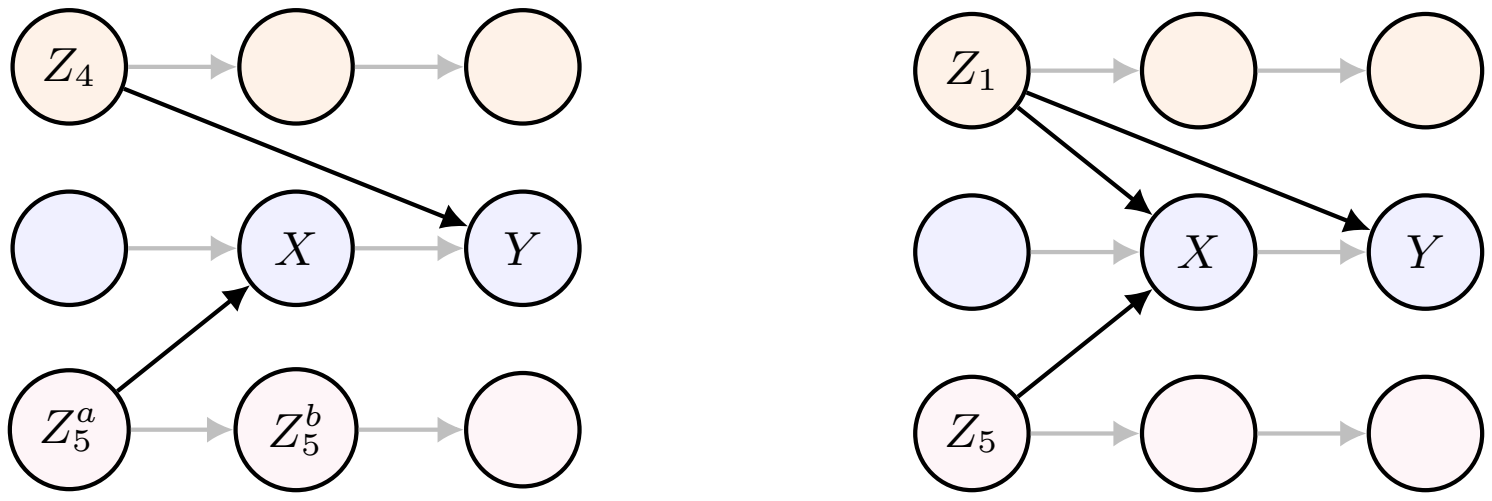
\begin{figure}[!h]
    \centering

\tikzset{
  A/.style={
    circle,
    draw=black,
    fill=A_color!30,
    thick,
    inner sep=3pt,
  },
  B/.style={
    circle,
    draw=black,
    fill=B_color!30,
    thick,
    inner sep=3pt,
  },
  C/.style={
    circle,
    draw=black,
    fill=C_color!30,
    thick,
    inner sep=3pt,
  }
}

\begin{tikzpicture}[
every node/.style={draw, minimum size=0.75cm},
every edge/.append style={thick, black, >={Latex[length=1.75mm,width=1.75mm]}},
scale=1,
font=\footnotesize
]


\node[A] (A1) at (0,4) {$Z_4$};
\node[A] (A2) at (1.5,4) {};
\node[A] (A3) at (3,4) {};

\node[B] (B1) at (0,2.8) {};
\node[B] (B2) at (1.5,2.8) {$X$};
\node[B] (B3) at (3,2.8) {$Y$};

\node[C] (C1) at (0,1.6) {$Z^a_5$};
\node[C] (C2) at (1.5,1.6) {$Z^b_5$};
\node[C] (C3) at (3,1.6) {};

\path (A1) edge[->,gray!50] (A2);
\path (A2) edge[->,gray!50] (A3);

\path (B1) edge[->,gray!50] (B2);
\path (B2) edge[->,gray!50] (B3);

\path (C1) edge[->,gray!50] (C2);
\path (C2) edge[->,gray!50] (C3);

\path (A1) edge[->] (B3);
\path (C1) edge[->] (B2);

\end{tikzpicture}
\hspace{2cm}
\begin{tikzpicture}[
every node/.style={draw, minimum size=0.75cm},
every edge/.append style={thick, black, >={Latex[length=1.75mm,width=1.75mm]}},
scale=1,
font=\footnotesize
]


\node[A] (A1) at (0,4) {$Z_1$};
\node[A] (A2) at (1.5,4) {};
\node[A] (A3) at (3,4) {};

\node[B] (B1) at (0,2.8) {};
\node[B] (B2) at (1.5,2.8) {$X$};
\node[B] (B3) at (3,2.8) {$Y$};

\node[C] (C1) at (0,1.6) {$Z_5$};
\node[C] (C2) at (1.5,1.6) {~};
\node[C] (C3) at (3,1.6) {};

\path (A1) edge[->,gray!50] (A2);
\path (A2) edge[->,gray!50] (A3);

\path (B1) edge[->,gray!50] (B2);
\path (B2) edge[->,gray!50] (B3);

\path (C1) edge[->,gray!50] (C2);
\path (C2) edge[->,gray!50] (C3);

\path (A1) edge[->] (B3);
\path (A1) edge[->] (B2);
\path (C1) edge[->] (B2);

\end{tikzpicture}

    \caption{Cases of $\z_5$. Assume no unobserved history. In the temporal DAG at left, $Z^a_5$ represents Case 1 (removed at Line 9) and $Z^b_5$ represents Case 3 (removed at Line 9). In the DAG at right, $Z_5$ represents Case 2. This variable would be removed from consideration at Line 14 if tested for candidacy first (before $Z_1$). As $\mathbf{S} = \{X\}$ under these conditions, $Z_5$ will pass through Line 9. As $Z_5$ has no parents in the observed graph and $Z_5 \not\ind Y$, it will pass through Line 10 as well. By Line 14, $Z_1$ will be in $\mathbf{S}'$, and $Z_5 \ind Y \mid \mathbf{S}' \setminus Z_5$.
    }
    \label{fig:z5}
\end{figure}

\begin{proof}
    We prove Theorem \ref{theorem:sound_complete_local} by cases, where each case is a causal partition wrt a given $\{X,Y\}$ pair. As stated in \S\ref{sec:itpd_naive}, the following partitions can arise in this problem setting: $\z_1$, $\z_{2 \in \nd(Y)}$, $\z_4$, $\z_5$, and $\z_8$. Of these, $\pa(Y) \subseteq \z_{1,4}$. Thus, we must show that (1) ITPD eliminates all of $\z_{5,8}$ and $\z_{2 \in \nd(Y)}$ and (2) ITPD retains \textit{all} and \textit{only} $\z_{1,4} \in \pa(Y)$ for any non-root $Y$. 

    Assume that $\mathbf{A}[V]=\pa(V)$ for every $V \in \mathbf{V}_{1:t}$ in all series. In our proof of Theorem \ref{theorem:sound_complete}, we will prove that this assumption holds. 
    
    \begin{enumerate}[label=\textit{Case \arabic*:}]
        \item \textit{Partition} $\z_8$. When $\mathbf{V}_1$ is not latently confounded, this partition is eliminated at Line 8 of Phase I by a marginal test. As only $\z_8$ is marginally independent of $Y$ when $X \to Y$, Line 8 is sound \citep{maasch2024local}. Line 7 is a sound gate for Line 8 as every member of $\mathbf{S}$ is an ancestor of $Y$, and an ancestor of $Y$ is not in $\z_8$ by definition. When $\mathbf{V}_1$ is confounded, $\z_8$ in the observed graph are removed at Line 14, maintaining exactness.
        \item \textit{Partition} $\z_{2 \in \nd(Y)}$ (Figure \ref{fig:z2}). By definition, this partition is disjoint with $\an(Y)$ \citep{maasch2024local}. Thus, $Y$ is a nondescendant of this set. When $\mathbf{V}_1$ is not latently confounded, this partition will be removed by the local Markov test at Line 10. Otherwise, it is removed at Line 14, maintaining exactness.
        \item \textit{Partition} $\z_5$. By definition, $\z_5 \cap \pa(Y) = \varnothing$. Thus, it must be removed by ITPD. 
        There are three cases of $\z_5$: (Case 1) ancestral to $Y$, where $\{X,Y\}$ are not confounded; (Case 2) ancestral to $Y$ where $\{X,Y\}$ are confounded; and (Case 3) not ancestral to $Y$ (proxy instruments). These three cases will all be removed by Lines 9, 10, and/or 14 (Figure \ref{fig:z5}). Per the arguments given in the following case, any non-parent that reaches Line 14 is removed.
        \item \textit{Partition} $\z_{1,4}$. All parents of $Y$ belong to this superpartition. First, we assume that $\mathbf{V}_1$ is not latently confounded. Variables in $\z_{1,4}$ that are not ancestral to $Y$ will be removed by the local Markov test, while ancestors will pass through to Line 13 (if not captured by Line 9). Additionally, all parents of $Y$ will be in $\mathbf{S}$ after Lines 8, 9 and 10 by the following logic. Any parent $Z$ of $Y$ is dependent on $Y$ given any conditioning set that contains neither $Y$ nor $Z$. This holds for the conditioning sets at Lines 8, 9 and 10, which lie in $\mathbf{V}_{1:t}$.
        When $\mathbf{V}_1$ is latently confounded, any variable in $\z_{1,4}$ that is not in $\pa(Y)$ will be removed by Line 14 in Phase II. To show correctness of Phase II, consider two facts: (1) $\pa(Y) \subseteq \mathbf{S}'$ at every pass, and (2) every other member of $\mathbf{S}'$ is a nondescendant of $Y$, so conditioning of these variables does not open paths between the candidate and $Y$. Thus, given $Z \in \an(Y), \not \in \pa(Y)$, conditioning on $\mathbf{S}'$ blocks the flow of association between $Y$ and $Z$, justifying the removal of $Z$ from $\mathbf{S}'$. This also covers candidates that are not ancestors of $Y$.
        Such conditioning cannot $d$-separate a direct cause from its child. Thus, Line 14 retains only the parents of $Y$. Any candidate that fails this test can be removed from the candidate set used for conditioning without jeopardizing the validity of Line 14 at subsequent iterations.
    \end{enumerate}
Therefore, the local parent discovery subroutine of ITPD is sound. Additionally, it is complete: as all parents are returned, and not a subset thereof, any parent discovery algorithm can only provide as much structural information or less than ITPD provides. 
\end{proof}


\begin{proof} We prove Theorem \ref{theorem:sound_complete} by induction. Let $\g = (\mathbf{V},\mathbf{E})$ be a full time DAG (Definition \ref{def:full_time_graph}) where $\mathbf{V} \in \mathbb{R}^{T \times N}$ is a stochastic process with discrete time index $t$ (Definition \ref{def:stochastic_process}). 

Let $Q(t)$ be the proposition that $\mathbf{A}[V^i_s]= \pa(V^i_s)$ for all series $i$ and all time steps $s \le t$. We prove by induction that $Q(t)$ holds for all $t \in \mathbb{N}$.

\textit{Base Case.} Let $Q(1)$ denote the base case where $t=1$. $Q(1)$ holds by Line 1 (with \ref{assumption:latent_variables}), as $V \in \mathbf{V}_1$ have no parents in the observed graph.

\textit{Induction Step.} Assume that $Q(j)$ is true for any $1 \leq j \leq k$ (induction hypothesis). 
We will show that $Q(k+1)$ is true:
\[
\forall k \in \mathbb{N}, Q(k) \Longrightarrow Q(k+1).
\]
If $Q(k)$ holds, all information that the iteration for $Y=V^n_{k+1}$ reads from $\mathbf{A}$ concerning $\mathbf{V}_{1:k}$ is correct. Thus, the assumption that $\mathbf{A}[V]=\pa(V)$ for $V \in \mathbf{V}_{1:k}$ in Theorem \ref{theorem:sound_complete_local} holds. 
When this assumption holds, the proof for Theorem \ref{theorem:sound_complete_local} dictates that the local parent discovery procedure for any $Y$ is sound and complete. Hence, $Q(k+1)$ is true.

We therefore conclude that $Q(t)$ is true for all $t \geq 1$.
\end{proof}

\subsection{Computational Complexity}
\label{appendix:complexity_itpd}

\begin{proof}
    We prove Theorem \ref{theorem:complexity} (polynomial computational complexity) as follows. Let stochastic process $\mathbf{V} \in \mathbb{R}^{T \times N}$ be a set of random variables that are time-indexed over a finite horizon by $t \in [0,\dots,T-1]$ for $T \in \mathbb{N}$. Let $\mathbf{V}_t \coloneqq \{V^n_t\}_{n=0}^{N-1}$ denote the $N$ variables at a given time step. Let $\mathbf{V}^n \coloneqq \{V^n_t\}_{t=0}^{T-1}$ denote the full time series for the $n^{th}$ variable. For convenience, time is zero-indexed to simplify calculations.

    At $t = 0$, all variables will be roots. Thus, zero CI tests are required. At $t = 1$, $O(N)$ CI tests will be performed for each of the $N$ variables in $\mathbf{V}_1$. As $t$ increases, we obtain the following summation:
    \begin{align}
        \sum_{t=0}^{T-1} N^2 t &= N^2 \sum_{t=0}^{T-1} t = N^2 \frac{(T-1)((T-1)+1)}{2} = \frac{N^2 (T^2 - T)}{2}.
    \end{align}
    This follows from the formula for summing the first $T-1$ natural numbers. Asymptotically, this yields $O(N^2 T^2)$. Thus, in the worst case, Algorithm \ref{alg:itpd} requires $O(N^2 T^2)$ CIT tests.
\end{proof}

\clearpage


\begin{algorithm}[t]
\caption{\textit{Parent Discovery on Leaves (PaDL) -- A subroutine of LD3 \citep{maasch2025local}} 
} \label{alg:padl}
    \begin{algorithmic}[1]
        \INPUT Pair of interest $\{X, Y\}$, variable set $\z$, CI test \cit, significance level $\alpha$, boolean flag \texttt{XisKnownParent}. Statistical dependence is determined by \cit at threshold $\alpha$.
        \OUTPUT Learned parent set $\widehat{\widehat{\pa}}(Y)$, partition label dictionary $\mathcal{D}$.
        \ASSUMPTIONS Sufficient conditions \ref{assumption:y_no_desc} and \ref{assumption:all_parents_y}; see \S\ref{sec:padl}.
        \vspace{3mm}
        \STATE $\z' \gets \z$
        \STATE $\mathcal{D}[Z] \gets$ `not identified' $\forall Z \in \z'$
        \FOR{$ Z \in \z'$}
            \STATE \textbf{if} $Z \ind X \land Z \ind Y$ \textbf{then} $\mathcal{D}[Z] \gets \widehat{\z}_8$ 
            \STATE \textbf{elif} $Z \nind Y \land Z \ind Y \mid X$ \textbf{then} $\mathcal{D}[Z] \gets \widehat{\z}_{5,7}$ 
            \STATE \textbf{elif} $Z \ind X \land Z \nind X \mid Y$ \textbf{then} $\mathcal{D}[Z] \gets \widehat{\z}_4$ 
        \ENDFOR
        \STATE $\z' \gets \z' \setminus \widehat{\z}_8 \cup \widehat{\z}_{5,7} \cup \widehat{\z}_4$
        \FOR{$ Z \in \z'$}
            \STATE \textbf{if} $Z \nind Y \mid X \cup \widehat{\z}_{4} \cup \{\z' \setminus Z\}$ 
            \textbf{then} $\mathcal{D}[Z] \gets \widehat{\z}_{1\in pa(Y)} \cup \widehat{\z}_{3\in pa(Y)}$  
        \ENDFOR
        \FOR{$ \widehat{Z}_4 \in \widehat{\z}_4$}
            \STATE \textbf{if} $\widehat{Z}_4 \nind Y \mid X \cup \widehat{\z}_{1\in pa(Y)} \cup \widehat{\z}_{3\in pa(Y)} \cup \{\widehat{\z}_4 \setminus \widehat{Z}_4\}$ \textbf{then} $\mathcal{D}[\widehat{Z}_4] \gets \widehat{\z}_{4\in pa(Y)}$ 
        \ENDFOR 
        \STATE $\widehat{\pa}(Y) \gets \widehat{\z}_{1\in pa(Y)} \cup \widehat{\z}_{3\in pa(Y)} \cup \widehat{\z}_{4\in pa(Y)}$ 
        \IF{not \texttt{XisKnownParent}}
            \STATE \textbf{if} $X \not\ind Y \mid \widehat{\pa}(Y)$ \textbf{then} $\widehat{\pa}(Y) \gets \widehat{\pa}(Y) \cup \{X\}$ 
        \ENDIF
        \RETURN $\widehat{\pa}(Y)$, $\mathcal{D}$
    \end{algorithmic}
\end{algorithm}

\begin{algorithm}[t]
\caption{\textit{Iterative Parent Discovery on Leaves (IPaDL)}
} \label{alg:itpd_naive}
    \begin{algorithmic}[1]
        \INPUT Multivariate time series $\mathbf{V} \in \mathbb{R}^{M \times T \times N}$ with $M$ replicate trajectories, CI test \texttt{CIT}, significance level $\alpha$,  boolean flag \texttt{XisKnownParent}. Statistical dependence is determined by \cit\, at threshold $\alpha$.
        \OUTPUT Adjacency list $\mathbf{A}$ where $\mathbf{A}[V^n_t] = \widehat{\pa}(V^n_t)$ for all $V^n_t \in \mathbf{V}$.
        \ASSUMPTIONS See \S\ref{sec:itpd_naive}.
        \vspace{3mm}
        \STATE $\mathbf{A} \gets \{V_1^i \, \mapsto \, \varnothing : i \in [1,\dots,N]\}$ \COMMENT{Initialize roots: parent set is empty.}
        \FOR{$t \in [1, \dots, T-1]$}
            \FOR{$n \in [1, \dots, N]$}
                \STATE $\widehat{\pa}(V^n_{t+1}), \mathcal{D} \gets$ \texttt{PaDL}($\{V^n_t, V^n_{t+1}\}$, $\mathbf{V}_{1:t} \setminus \{V^n_t, V^n_{t+1}\}$, \texttt{CIT}, $\alpha$, \texttt{XisKnownParent})
                \STATE $\mathbf{A}[V^n_{t+1}] \gets \widehat{\pa}(V^n_{t+1})$
            \ENDFOR
        \ENDFOR
        \RETURN $\mathbf{A}$
    \end{algorithmic}
\end{algorithm}

\section{Proofs: IPaDL (Algorithm \ref{alg:itpd_naive})}
\label{appendix:ipadl}

\subsection{Parent Discovery on Leaves}

\label{sec:padl}We briefly summarize PaDL (Algorithm \ref{alg:padl}). We refer the reader to \citet{maasch2025local} for proofs. 

\paragraph{Intuition} See Algorithm \ref{alg:padl} for full pseudocode. PaDL takes in cause $X$, effect $Y$, set $\z$, a CI test of choice, and a significance threshold $\alpha$. A sequence of CI tests is performed to causally partition $\z$ with respect to $\{X,Y\}$. PaDL returns \textit{partition labels}, not a causal graph. When sufficient conditions are satisfied, this partitioning procedure is asymptotically guaranteed to return the parent set of $Y$.

\paragraph{Sufficient Conditions} As in LD3 \citep{maasch2025local}, theoretical guarantees on correctness rely on acyclicity, causal Markov, faithfulness, and the following sufficient conditions.
\begin{enumerate}[label=A\arabic*,leftmargin=\widthof{A1}+\labelsep,start=4] 
    \item \textit{No observed descendants of $Y$: $\de(Y) \cap \z = \varnothing$.} \label{assumption:y_no_desc}
    \item \textit{All of $\pa(Y)$ is observed in $\z$: $\pa(Y) \subseteq \z$.} \label{assumption:all_parents_y}
\end{enumerate}
\ref{assumption:y_no_desc} is satisfied when $Y$ is a leaf (e.g., $Y$ is terminal in the time series).  Full causal sufficiency is not required, as latent variables that are not in $\pa(Y)$ do not impact correctness \citep{maasch2025local}. 
This procedure makes no parametric assumptions about the distributional forms of random variables nor the functional forms of $\mathcal{F}$, though the user-selected CI test might.

\paragraph{Computational Complexity} PaDL performs $O(d)$ CI tests wrt $d$  variables \citep{maasch2025local}. 

\subsection{A Naive Extension for Time Series}
\label{sec:itpd_naive}

Next, we present IPaDL: a simplistic adaptation of PaDL to the TiMINo setting for \ref{setting:single} and \ref{setting:multi}. Our objective is to learn the parent set for every variable, yielding an iterative local-to-global discovery procedure. Though asymptotically guaranteed to return the correct graph under sufficient conditions, PaDL was designed for a setting with different structural priors: namely, no time-indexing and unknown relation between $X$ and $Y$. Thus, naively applying PaDL to each variable results in extraneous CI tests and unnecessarily large conditioning sets. IPaDL therefore serves as a proof-of-concept that iterative local parent discovery is a viable strategy for temporal graph learning, but thoughtful extensions are required for efficiency.

\paragraph{Assumptions} IPaDL assumes a TiMINo that satisfies the following conditions: acyclicity, jointly independent exogenous noise terms, causal Markov, faithfulness, autocorrelation, and no contemporaneous links. Causal stationarity is not required, though a window graph of maximum lag $\tau_{\max}$ can be assumed if desired. Though PaDL allows for latent variables that are not in $\pa(Y)$, the iterative application of PaDL to all variables requires that all variables are observed. We make no parametric assumptions, though the user-selected CI test may. We assume that all variables are measured at the same time steps $[1, \dots, T]$. In addition to \ref{assumption:y_no_desc} and \ref{assumption:all_parents_y}, IPaDL assumes \ref{assumption:autocorrelation} and \ref{assumption:no_contem}. PaDL is well suited for SCMs with autocorrelation and no contemporaneous links: 
\begin{enumerate}
    \item Each cause-effect pair can simply be set to $\{V^n_t, V^n_{t+1}\}$ for $t \in [1, \dots, T-1]$.
    \item The known temporal ordering allows the user to easily satisfy \ref{assumption:y_no_desc}.  When $V^i_{t}$ is the effect variable of interest, the user can simply remove all other variables in $\mathbf{V}^i_{t+1:T}$ and $\mathbf{V}^j_{t:T}$ ($i \neq j$) from $\z$.
\end{enumerate}

\paragraph{Relevant Partitions} This problem setting admits the existence of 5 partitions wrt each cause-effect pair of interest: $\z_1$, $\z_{2 \in \nd(Y)}$, $\z_4$, $\z_5$, and $\z_8$. Because $Y$ is a leaf, $\z_6 = \varnothing$ and $\z_{2 \in de(Y)} = \varnothing$. Because we take $X$ and $Y$ to be autocorrelated variables at $t$ and $t+1$ ($V_t^n \to V_{t+1}^n$) and contemporaneous links are not permitted, there are no descendants of $X$ to consider ($\z_{3,7} = \varnothing$). 
See Figures  \ref{fig:partition_examples} and \ref{fig:z2} for examples of causal partitions in full time graphs.

\paragraph{Naive Implementation} It is sufficient to simply run PaDL iteratively for each autocorrelated pair $V^n_t \to V^n_{t+1}$. The predicted parent set for $V^n_{t+1}$ then becomes a column in the global adjacency matrix. We refer to this iterative application of PaDL as IPaDL (Algorithm \ref{alg:itpd_naive}). This implementation tests all autocorrelated pairs forward in time from $t = 1$ to $t = T-1$. We show that this naive implementation results in an acyclic graph under infinite and finite data (Theorem \ref{theorem:cycles}), and that the resulting graph is asymptotically correct (Theorem \ref{theorem:naive_correct}). 

\begin{theorem}[Iterative application of PaDL returns an acyclic adjacency matrix] \label{theorem:cycles}
    Let $\g$ be a time series causal graph representing a TiMINo that satisfies Assumptions \ref{assumption:autocorrelation} and \ref{assumption:no_contem}. Applying PaDL to every node in $\mathcal{G}$ does not induce cycles in the learned global graph, in both the asymptotic and finite sample regime.
\end{theorem}
\textit{Intuition.} This result follows from the fact that variables are time-indexed and all true edges go forward in time.  PaDL uses temporal ordering as a prior to exclude candidate parents, forming a hard constraint that is not impacted by finite sample artifacts.

We prove Theorem \ref{theorem:cycles}, which follows from the fact that variables are time-indexed and all true edges go forward in time.  PaDL uses temporal ordering as a prior to exclude candidate parents, forming a hard constraint that is not impacted by finite sample artifacts.
\begin{proof}
     Given a time series DAG with no contemporaneous links, all edges go forward in time. A cycle could arise if a discovery algorithm erroneously infers edges going back in time: $V^i_t \to V^j_{t-k}$ for $k > 0$. For input $\{V^n_t, V^n_{t+1}\}$, PaDL only considers variables at $[1, \dots, t]$ as candidate parents. Thus, it is impossible for the iterative application of PaDL to infer edges going back in time, and the adjacency matrix will be acyclic.
\end{proof}

\begin{theorem}[Iterative application of PaDL returns the correct adjacency matrix] \label{theorem:naive_correct}
    Given an oracle independence test, applying PaDL to every node in $\mathcal{G}$ returns the correct adjacency matrix. 
\end{theorem}
\textit{Intuition.} As each column of the adjacency matrix will be asymptotically correct \citep{maasch2025local}, 
then the global matrix will also be asymptotically correct.

We prove Theorem \ref{theorem:naive_correct}, which states that the adjacency matrix returned by IPaDL is asymptotically correct. 

\begin{proof}
    As previously shown \citep{maasch2025local}, PaDL returns the asymptotically correct parent set for a node under sufficient conditions. The full adjacency matrix is comprised of columns representing the parent set of each node. Thus, identifying the true parent set of each node is sufficient to identify the full graph. As each column of the matrix is asymptotically correct, then the full matrix is also asymptotically correct.
\end{proof}

\begin{remark}
    Theorem \ref{theorem:naive_correct} addresses the infinite data setting. In finite samples, it is possible that conflicting conclusions can arise between the columns of the predicted adjacency matrix. However, as iterative application of PaDL does not induce cycles even in finite data (Theorem \ref{theorem:cycles}), these conflicts will not take the form of acyclicity violations.
\end{remark}

\paragraph{Computational Complexity} IPaDL performs $O(N^2T^2)$ CI tests. We can reduce this to $O(N^2\tau_{\max}^2)$ if we operate in setting \ref{setting:single} for a window graph of order $\tau_{\max} \ll T$. The complexity of IPaDL wrt total CI tests is the same as given for ITPD, and shares the same proof (Appendix \ref{appendix:complexity_itpd}).

\paragraph{Known Limitations} While IPaDL is sound and complete, 
redundant CI tests are performed and conditioning set are unnecessarily large. In finite samples, poor sample efficiency due to high-order CI tests can lead to incorrect inferences and error propagation. Thus, we relegate IPaDL to the role of a naive baseline.

\begin{figure}[t]
    \centering

\tikzset{
  X/.style={
    circle,
    draw=parents,
    fill=X_color!20,
    ultra thick,
    inner sep=2.5pt,
    text=black
  },
  Y/.style={
    circle,
    draw=black,
    fill=Y_color!20,
    ultra thick,
    inner sep=2.5pt,
    text=black
  },
  Z1/.style={
    circle,
    draw=black,
    fill=Z1_color!40,
    thick,
    inner sep=2.5pt,
    text=black
  },
  Z4/.style={
    circle,
    draw=black,
    fill=Z4_color!20,
    thick,
    inner sep=2.5pt,
    text=black
  },
  Z5/.style={
    circle,
    draw=black,
    fill=Z5_color!20,
    thick,
    inner sep=2.5pt,
    text=black
  },
  Z6/.style={
    circle,
    draw=black,
    fill=Z6_color!20,
    thick,
    inner sep=2.5pt,
    text=black
  },
  Z7/.style={
    circle,
    draw=black,
    fill=Z7_color!20,
    thick,
    inner sep=2.5pt,
    text=black
  },
  Z8/.style={
    circle,
    draw=black,
    fill=Z8_color!20,
    thick,
    inner sep=2.5pt,
    text=black
  }
}

\begin{tikzpicture}[
every edge/.append style={thick, black, >={Latex[length=1.75mm,width=1.75mm]}},
scale=1,
font=\footnotesize
]


\node[Z4] (A1) at (0,4) {A$_1$};
\node[Z4,draw=parents,ultra thick] (A2) at (1.5,4) {A$_2$};
\node[Z4] (A3) at (3,4) {A$_3$};
\node[Z4] (A4) at (4.5,4) {A$_4$};
\node[Z4] (A5) at (6,4) {A$_5$};

\node[Z1] (B1) at (0,2.8) {B$_1$};
\node[Z1] (B2) at (1.5,2.8) {B$_2$};
\node[Z1,draw=parents,ultra thick] (B3) at (3,2.8) {B$_3$};
\node[Z1] (B4) at (4.5,2.8) {B$_4$};
\node[Z1] (B5) at (6,2.8) {B$_5$};

\node[Z5] (C1) at (0,1.6) {C$_1$};
\node[Z5] (C2) at (1.5,1.6) {C$_2$};
\node[X] (C3) at (3,1.6) {C$_3$};
\node[Y] (C4) at (4.5,1.6) {C$_4$};
\node[Z6] (C5) at (6,1.6) {C$_5$};

\node[Z8] (D1) at (0,0.4) {D$_1$};
\node[Z8] (D2) at (1.5,0.4) {D$_2$};
\node[Z8] (D3) at (3,0.4) {D$_3$};
\node[Z7] (D4) at (4.5,0.4) {D$_4$};
\node[Z7] (D5) at (6,0.4) {D$_5$};


\path (A1) edge[->,gray!70] (A2);
\path (A2) edge[->,gray!70] (A3);
\path (A3) edge[->,gray!70] (A4);
\path (A4) edge[->,gray!70] (A5);

\path (B1) edge[->,gray!70] (B2);
\path (B2) edge[->,gray!70] (B3);
\path (B3) edge[->,gray!70] (B4);
\path (B4) edge[->,gray!70] (B5);

\path (C1) edge[->,gray!70] (C2);
\path (C2) edge[->,gray!70] (C3);
\path (C3) edge[->,parents] (C4);
\path (C4) edge[->,gray!70] (C5);

\path (D1) edge[->,gray!70] (D2);
\path (D2) edge[->,gray!70] (D3);
\path (D3) edge[->,gray!70] (D4);
\path (D4) edge[->,gray!70] (D5);


\path (A2) edge[->,bend left=25,parents] (C4);

\path (B2) edge[->] (C3);
\path (B3) edge[->,parents] (C4);

\path (C3) edge[->] (D4);




\end{tikzpicture}

    \caption{A full time graph representing a stochastic process over finite horizon $T=5$, featuring autocorrelation and causal nonstationarity. Our cause-effect pair of interest is $\{C_3,C_4\}$. Relative to this pair, we illustrate partitions $\z_1 = \{{\color{Z1_color}B_{1:5}}\}$, $\z_4 = \{{\color{Z4_color}A_{1:5}}\}$, $\z_5 = \{{\color{Z5_color}C_{1:2}}\}$, $\z_6 = \{{\color{Z6_color}C_5}\}$, $\z_7 = \{{\color{Z7_color}D_{4:5}}\}$, and $\z_8 = \{{\color{Z8_color}D_{1:3}}\}$. Causal edges that induce autocorrelations are gray. Adjacencies to effect variable $C_4$ are {\color{parents}red}. All other causal edges are black. Given $\{C_3,C_4\}$, ITPD is expected to return $\{A_2,B_3\}$. Unioned with known parent $C_3$, this constitutes $\pa(C_4)$. Note that $\z_{6,7}$ will not be relevant when ITPD infers $\pa(C_4)$. See Figure \ref{fig:z2} for an example of $\z_{2 \not\in de(Y)}$.}
    \label{fig:partition_examples}
\end{figure}
\definecolor{A_color}{HTML}{E9692C}
\definecolor{B_color}{HTML}{F4BBFF}
\definecolor{C_color}{HTML}{96C8A2}

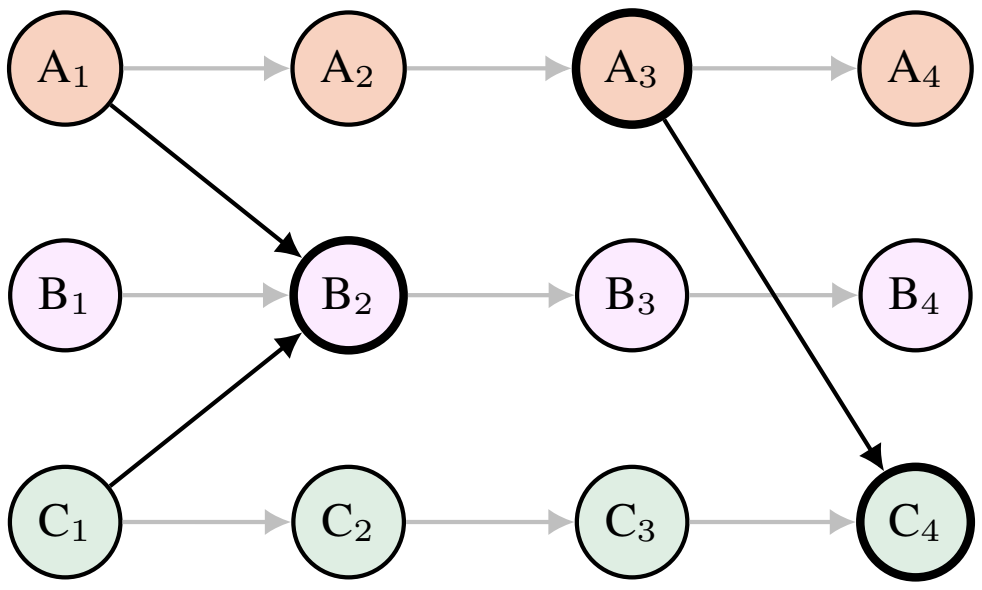
\begin{figure}[t]
    \centering

\tikzset{
  A/.style={
    circle,
    draw=black,
    fill=A_color!30,
    thick,
    inner sep=3pt,
  },
  B/.style={
    circle,
    draw=black,
    fill=B_color!30,
    thick,
    inner sep=3pt,
  },
  C/.style={
    circle,
    draw=black,
    fill=C_color!30,
    thick,
    inner sep=3pt,
  }
}

\begin{tikzpicture}[
every edge/.append style={thick, black, >={Latex[length=1.75mm,width=1.75mm]}},
scale=1.25,
font=\footnotesize
]


\node[A] (A1) at (0,4) {A$_1$};
\node[A] (A2) at (1.5,4) {A$_2$};
\node[A, ultra thick] (A3) at (3,4) {A$_3$};
\node[A] (A4) at (4.5,4) {A$_4$};

\node[B] (B1) at (0,2.8) {B$_1$};
\node[B,ultra thick] (B2) at (1.5,2.8) {B$_2$};
\node[B] (B3) at (3,2.8) {B$_3$};
\node[B] (B4) at (4.5,2.8) {B$_4$};

\node[C] (C1) at (0,1.6) {C$_1$};
\node[C] (C2) at (1.5,1.6) {C$_2$};
\node[C] (C3) at (3,1.6) {C$_3$};
\node[C, ultra thick] (C4) at (4.5,1.6) {C$_4$};

\path (A1) edge[->,gray!50] (A2);
\path (A2) edge[->,gray!50] (A3);
\path (A3) edge[->,gray!50] (A4);

\path (B1) edge[->,gray!50] (B2);
\path (B2) edge[->,gray!50] (B3);
\path (B3) edge[->,gray!50] (B4);

\path (C1) edge[->,gray!50] (C2);
\path (C2) edge[->,gray!50] (C3);
\path (C3) edge[->,gray!50] (C4);

\path (A1) edge[->] (B2);
\path (C1) edge[->] (B2);
\path (A3) edge[->] (C4);

\end{tikzpicture}

    \caption{Time series causal graph where node $B_2$ is an example of $\z_{2 \not \in \de(Y)}$ wrt cause-effect pair $\{A_3,C_4\}$. This example is an \textit{M-collider}, the culprit behind M-bias \citep{liu2012implications,ding2015adjust}. Causal edges that induce autocorrelations are gray. All other causal edges are black.}
    \label{fig:z2}
\end{figure}

\clearpage



\section{Autoformalization in Lean}
\label{appendix:autoform}
\sloppy

\subsection{Overview}
The formalization is in the folder \texttt{lean/} of the code repository. It uses Lean 4.33.1 and its core library only, with no Mathlib or other dependency. The code contains 7 modules and 61 theorem declarations, including supporting lemmas. Every proof is complete with no \texttt{sorry}, and the library declares no axiom. The development proves that Algorithm~\ref{alg:itpd} returns the exact parent set $\mathbf{A}[Y] = \pa(Y)$ for every target $Y$ whose parents are all observed, when the conditional independence (CI) answers match the separation statements of the graph. The graph may have any number $H \ge 0$ of latent time steps before the first observed one. No initial condition is assumed, so variables at the first observed time step may have latent parents. The theorems for Algorithm~\ref{alg:itpd} are instances of general theorems that hold for every order of the candidates in Phase I (Line 6), every order in Phase II (Line 13), and every order of the series (Line 3).

\subsection{Formal Model \& Assumptions}
Vertices are pairs of a time step and a series, and a graph is a directed edge relation. The structure \texttt{LatentDAG N T H} represents $H$ latent time steps followed by $T$ observed ones. \ref{assumption:autocorrelation} (lag-1 self edge) is the field \texttt{LatentDAG.lag1} on the observed time steps. \ref{assumption:no_contem} (no edge inside a time step) is \texttt{LatentDAG.forward}, which requires every edge to point strictly forward in time. \ref{assumption:latent_variables} enters as the hypothesis \texttt{ParentsObserved G t n} for each target, stating that all parents of the target are observed. With links of lag at most $L$, every target after time step $L$ satisfies this; with lag-1 links, every target does.

Separation is defined on the moral graph of the ancestral set via \texttt{MSep.msep}. The predicate \texttt{LatentOracle G ci} states that Boolean CI answers equal this separation in the full graph, including latent time steps. The proof uses only two properties of the CI answers for a target, bundled in \texttt{TargetHyp}: a parent of $Y$ is reported dependent on $Y$ given every subset of $\mathbf{V}_{1:t}$ not containing it; and if $\pa(Y) \subseteq C \subseteq \mathbf{V}_{1:t}$, every $Z \in \mathbf{V}_{1:t} \setminus C$ that is not a parent of $Y$ is reported independent of $Y$ given $C$. The lemma \texttt{latent\_targetHyp} derives both from \texttt{LatentOracle}.

\subsection{Correspondence with the Paper}
All names below reside in the namespace \texttt{ITPD}, and in the file \texttt{ITPD/Paper.lean} unless stated otherwise.
\begin{itemize}
\item \texttt{itpdPaper} defines Algorithm~\ref{alg:itpd} in terms of \texttt{ITPD/Algorithm.lean}. Reflexivity lemmas (\texttt{rfl}) verify the tests of one step: \texttt{itpdPaper\_lines\_7\_10} covers Lines 7 to 10 (gating on learned ancestors of $\mathbf{S}$, testing given $\emptyset$, given $\mathbf{S}$, and given learned parents of $Z$), and \texttt{itpdPaper\_line\_11} covers $\widehat{\pa}(Y) \leftarrow \{X\}$.
\item For the local result (Theorem~\ref{theorem:sound_complete_local}), \texttt{itpd\_exact} uses abstract CI answers with \texttt{TargetHyp}, while \texttt{itpd\_exact\_latent} uses separation answers in a graph with latent time steps and a target satisfying \texttt{ParentsObserved}.
\item For the global result (Theorem~\ref{theorem:sound_complete}), \texttt{itpd\_exact\_latent\_lag1} covers every target under lag-1 links, and \texttt{itpd\_exact\_latent\_lag} covers every target after time step $L$ under links of lag at most $L$.
\item \texttt{itpd\_exact\_nonvacuous} shows the hypotheses hold on a graph where every variable of the first observed time step has latent parents.
\item The general ordering theorems are \texttt{itpdS\_correct} (\texttt{ITPD/Exact.lean}) and \texttt{itpdS\_correct\_latent} (\texttt{ITPD/Latent.lean}).
\end{itemize}

\subsection{Proof Audit, Reproducibility \& Scope}
Running \texttt{\#print axioms} on the five main theorems (\texttt{itpd\_exact}, \texttt{itpd\_exact\_latent}, \texttt{itpd\_exact\_latent\_lag}, \texttt{itpd\_exact\_latent\_lag1}, \texttt{itpd\_exact\_nonvacuous}) reports only \texttt{propext}, \texttt{Classical.choice}, and \texttt{Quot.sound}, the three standard Lean 4 axioms. To reproduce, install Lean via \texttt{elan}; \texttt{lean-toolchain} pins Lean 4.33.1. Run \texttt{lake build} in \texttt{lean/}. Because there are no package dependencies, the build downloads nothing after toolchain installation.

The formalization omits significance levels and test statistics, modeling a CI test by its Boolean answer. It also omits the link from data to CI answers (Markov and faithfulness properties for probability distributions), the equivalence of moral-graph separation with path-based d-separation, and the Python implementation. Finally, the theorems hold for every learned adjacency list $\mathbf{A}$ whose entries contain only earlier variables, so they do not depend on how the learned ancestors at Line 7 are obtained.

\fussy

\section{Extended Empirics}
\label{appendix:results}

\subsection{Experimental Design}
\label{appendix:experimental_design}

\textbf{Baselines} We compared ITPD and IPaDL against the following: 
\begin{itemize}
    \item \textit{Full conditioning}. For each $V^n_t$ of interest and each candidate parent $Z^i_k$ for $k < t$, we test for independence by $Z^i_k \ind V^n_t \mid \mathbf{V}_{1:t-1} \setminus \{Z^i_k\}$.
    \item \textit{IAMB} \citep{tsamardinos2003algorithms}. The Incremental Association Markov Blanket algorithm is a constraint-based local discovery algorithm for MB learning, which we have adapted to our setting by providing the known time order. This approach is analogous to IPaDL, in the sense that we iteratively apply a local learning procedure that is not specifically tailored to the current setting.
    \item \textit{PCMCI} \citep{runge2018causal}. 
    PCMCI is a constraint-based discovery algorithm for window graphs, which assumes causal stationarity and no contemporaneous links. Thus, it is expected to fail when the true graph changes over time. 
    \item \textit{DF-PC} \citep{kim2026deduce}. Deduce-First PC is a constraint-based algorithm that uses logical deduction with graph axioms to avoid higher-order CI tests. DF-PC is not explicitly designed to handle temporal data. We combine DF-PC with the known time order.
    \item \textit{DYNOTEARS} \citep{pamfil2020dynotears}. DYNOTEARS is score-based method that minimizes a penalized loss subject to an acyclicity constraint. The implementation used assumes linear causal functions. We select a weight threshold of 0.05.
    \item \textit{Lasso regression} \citep{tibshirani1996regression}. This noncausal baseline assumes linear functions.
    \item \textit{CD-NOD} \citep{huang2020causal}. CD-NOD is a constraint-based method that adds a time or domain index to the observed variables as a surrogate variable. We use the implementation from Andrey~\citep{andrey}.
\end{itemize}



\paragraph{Choice of $\gamma$ and $\kappa$}
We set $\gamma = 0.2$ and $\kappa = 50$. Both were chosen by F1 on simulated linear-Gaussian graphs and used unchanged in all other experiments; neither is derived from theory. Among the fixed Phase I levels $0.01$, $0.05$, $0.1$, and $0.2$, the level $0.2$ had the highest F1 at $M = 200$ and $M = 500$. The term $\kappa/m$ kept this F1 (within $0.001$ in all $16$ settings) and limited the size of $\mathbf{S}$ on larger graphs: at $N = 100$ and $T = 20$, the largest conditioning set was $27$ to $37$, against $134$ to $156$ with the fixed level $0.2$.

\paragraph{Dependence on the Number of Observations}
Neither level uses the number of observations $M$. An alternative for Phase II that does is the extended-BIC level, $\alpha_2' = P(\chi^2_1 > \log(M m))$ \citep{chen2008extended}. Against $\alpha_2 = \alpha/m$ on held-out settings (Table~\ref{tab:extended_bic}), it changed the F1 of ITPD by $+0.136$ at $M = 50$ (one setting), $+0.034$ at $M = 200$, and $-0.012$ at $M = 500$ and at $M \ge 2000$. It is an option when observations are few; ITPD keeps $\alpha/m$ as the default.

\paragraph{Compute Resources} Each experimental run used one CPU process on one core in a batch job on a university cluster.
We report wall-clock seconds, which depend on the CI test implementation and node load.


\subsection{Extended Results}

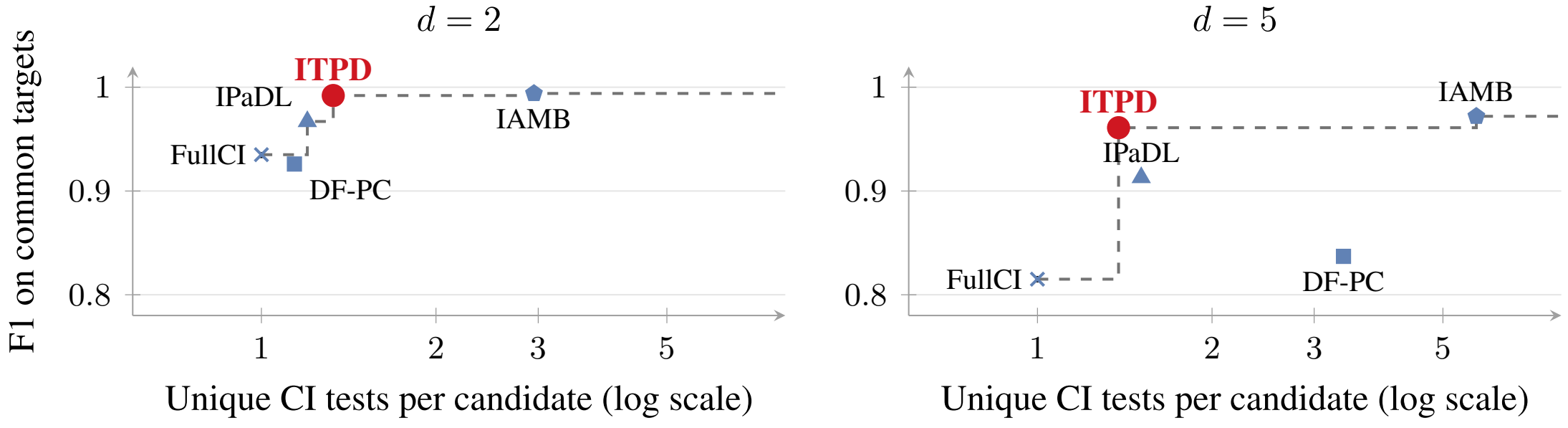
\begin{figure}[h] 
    \centering
    
\begin{tikzpicture}[scale=0.95]
\begin{axis}[
    name=frontA, 
    resaxis, width=0.52\linewidth, height=4.0cm,
    xmode=log, xmin=0.6, xmax=8, ymin=0.78, ymax=1.02,
    xtick={1,2,3,5}, log ticks with fixed point, ytick={0.8,0.9,1},
    xlabel={Unique CI tests per candidate (log scale)}, ylabel={F1 on common targets},
    title={$d = 2$}, clip=false
]
\addplot[black!55, thick, dashed] coordinates {(1.00,0.935) (1.20,0.935) (1.20,0.967) (1.33,0.967) (1.33,0.992) (2.95,0.992) (2.95,0.994) (8,0.994)};
\addplot[only marks, mark=*, mark size=3pt, parents] coordinates {(1.33,0.992)};
\draw[black, line width=0.7pt] (axis cs:1.3336,0.99166) -- (axis cs:1.3362,0.99166) (axis cs:1.3349,0.99052) -- (axis cs:1.3349,0.99271);
\node[anchor=south, inner sep=4pt, font=\footnotesize\bfseries, text=parents] at (axis cs:1.33,0.992) {ITPD};
\addplot[mIAMB, only marks,glaucous] coordinates {(2.95,0.994)};
\draw[black, line width=0.7pt] (axis cs:2.9159,0.99437) -- (axis cs:2.9901,0.99437) (axis cs:2.9537,0.99301) -- (axis cs:2.9537,0.99566);
\node[anchor=north, inner sep=4pt, font=\scriptsize] at (axis cs:2.95,0.994) {IAMB};
\addplot[mIPaDL, only marks,glaucous] coordinates {(1.20,0.967)};
\draw[black, line width=0.7pt] (axis cs:1.1933,0.9668) -- (axis cs:1.1969,0.9668) (axis cs:1.1951,0.96451) -- (axis cs:1.1951,0.96917);
\node[anchor=south east, inner sep=4pt, font=\scriptsize] at (axis cs:1.20,0.967) {IPaDL};
\addplot[mFullCI, only marks,glaucous] coordinates {(1.00,0.935)};
\draw[black, line width=0.7pt] (axis cs:1,0.93475) -- (axis cs:1,0.93475) (axis cs:1,0.93168) -- (axis cs:1,0.93762);
\node[anchor=east, inner sep=4pt, font=\scriptsize] at (axis cs:1.00,0.935) {FullCI};
\addplot[mDFPC, only marks,glaucous] coordinates {(1.14,0.926)};
\draw[black, line width=0.7pt] (axis cs:1.1363,0.92565) -- (axis cs:1.1433,0.92565) (axis cs:1.1398,0.92223) -- (axis cs:1.1398,0.92912);
\node[anchor=north west, inner sep=4pt, font=\scriptsize] at (axis cs:1.14,0.926) {DF-PC};
\end{axis}
\begin{axis}[
     at={(frontA.south east)}, anchor=south west, xshift=1.2cm,
    resaxis, width=0.52\linewidth, height=4.0cm,
    xmode=log, xmin=0.6, xmax=8, ymin=0.78, ymax=1.02,
    xtick={1,2,3,5}, log ticks with fixed point, ytick={0.8,0.9,1},
    xlabel={Unique CI tests per candidate (log scale)}, 
    title={$d = 5$}, clip=false
]
\addplot[black!55, thick, dashed] coordinates {(1.00,0.815) (1.38,0.815) (1.38,0.961) (5.71,0.961) (5.71,0.972) (8,0.972)};
\addplot[only marks, mark=*, mark size=3pt, parents] coordinates {(1.38,0.961)};
\draw[black, line width=0.7pt] (axis cs:1.3739,0.96059) -- (axis cs:1.3776,0.96059) (axis cs:1.3757,0.95885) -- (axis cs:1.3757,0.96229);
\node[anchor=south, inner sep=4pt, font=\footnotesize\bfseries, text=parents] at (axis cs:1.38,0.961) {ITPD};
\addplot[mIAMB, only marks,glaucous] coordinates {(5.71,0.972)};
\draw[black, line width=0.7pt] (axis cs:5.6637,0.97226) -- (axis cs:5.754,0.97226) (axis cs:5.7083,0.97104) -- (axis cs:5.7083,0.97342);
\node[anchor=south, inner sep=4pt, font=\scriptsize] at (axis cs:5.71,0.972) {IAMB};
\addplot[mIPaDL, only marks,glaucous] coordinates {(1.51,0.913)};
\draw[black, line width=0.7pt] (axis cs:1.5038,0.91256) -- (axis cs:1.514,0.91256) (axis cs:1.5086,0.9096) -- (axis cs:1.5086,0.91556);
\node[anchor=south, inner sep=4pt, font=\scriptsize] at (axis cs:1.51,0.913) {IPaDL};
\addplot[mFullCI, only marks,glaucous] coordinates {(1.00,0.815)};
\draw[black, line width=0.7pt] (axis cs:1,0.81495) -- (axis cs:1,0.81495) (axis cs:1,0.81184) -- (axis cs:1,0.81809);
\node[anchor=east, inner sep=4pt, font=\scriptsize] at (axis cs:1.00,0.815) {FullCI};
\addplot[mDFPC, only marks,glaucous] coordinates {(3.37,0.837)};
\draw[black, line width=0.7pt] (axis cs:3.315,0.83731) -- (axis cs:3.427,0.83731) (axis cs:3.3698,0.83455) -- (axis cs:3.3698,0.84019);
\node[anchor=north, inner sep=4pt, font=\scriptsize] at (axis cs:3.37,0.837) {DF-PC};
\end{axis}
\end{tikzpicture}
    \caption{ITPD lies on the Pareto frontier (dashed line) wrt mean F1 and and unique CI tests per candidate. $M = 500$ and $d \in \{2,5\}$. Each point is a mean over 20 graphs.
    }
    \label{fig:frontier_tests}
\end{figure}


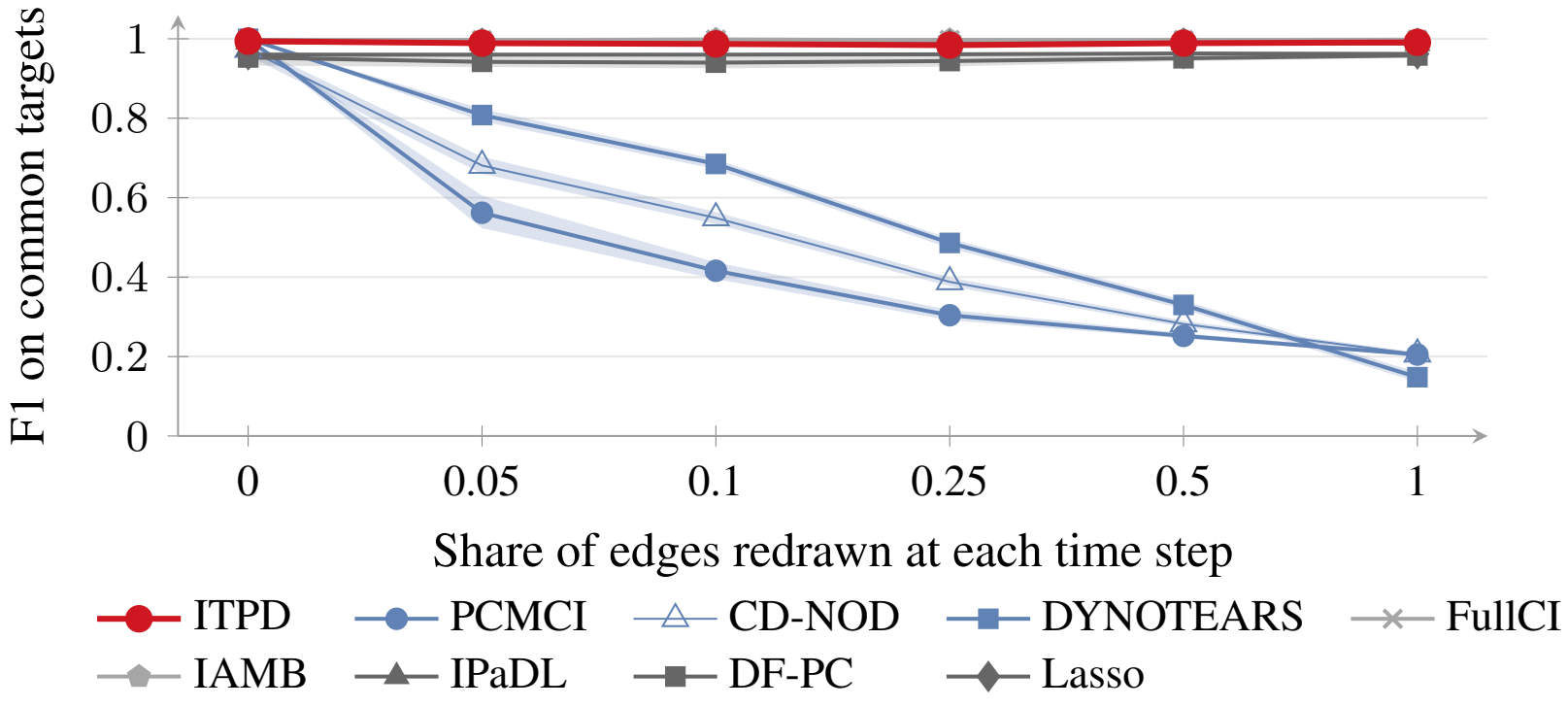
\begin{figure}[!htb] 
    \centering
\begin{tikzpicture}[scale=0.95]
\begin{axis}[
    resaxis, width=0.72\linewidth, height=4.6cm,
    xmin=-0.3, xmax=5.3, ymin=0, ymax=1.06,
    xtick={0,1,2,3,4,5}, xticklabels={0,0.05,0.1,0.25,0.5,1}, ytick={0,0.2,0.4,0.6,0.8,1},
    xlabel={Share of edges redrawn at each time step}, ylabel={F1 on common targets},
    legend columns=5,
    legend style={at={(0.5,-0.34)}, anchor=north}
]
\addlegendimage{mITPD} \addlegendentry{ITPD}
\addplot[fill=glaucous, fill opacity=0.22, draw=none, no markers, forget plot] coordinates {(0,1) (1,0.60515) (2,0.43725) (3,0.31744) (4,0.25906) (5,0.20921) (5,0.20128) (4,0.24536) (3,0.29157) (2,0.39479) (1,0.52317) (0,1)} -- cycle;
\addplot[fill=glaucous, fill opacity=0.22, draw=none, no markers, forget plot] coordinates {(0,0.98666) (1,0.70283) (2,0.56386) (3,0.39946) (4,0.28993) (5,0.21068) (5,0.20058) (4,0.2744) (3,0.37683) (2,0.53361) (1,0.66006) (0,0.95747)} -- cycle;
\addplot[fill=glaucous, fill opacity=0.22, draw=none, no markers, forget plot] coordinates {(0,1) (1,0.82343) (2,0.69685) (3,0.4971) (4,0.3415) (5,0.15999) (5,0.13569) (4,0.31883) (3,0.47305) (2,0.67213) (1,0.79213) (0,1)} -- cycle;
\addplot[fill=black!55, fill opacity=0.22, draw=none, no markers, forget plot] coordinates {(0,0.99778) (1,0.99782) (2,0.99801) (3,0.99804) (4,0.99762) (5,0.99804) (5,0.99704) (4,0.99621) (3,0.99676) (2,0.99712) (1,0.99674) (0,0.99661)} -- cycle;
\addplot[fill=black!55, fill opacity=0.22, draw=none, no markers, forget plot] coordinates {(0,0.99801) (1,0.99784) (2,0.99834) (3,0.99793) (4,0.99787) (5,0.99758) (5,0.99641) (4,0.99679) (3,0.99683) (2,0.9971) (1,0.9968) (0,0.99663)} -- cycle;
\addplot[fill=black!55, fill opacity=0.22, draw=none, no markers, forget plot] coordinates {(0,0.99572) (1,0.99229) (2,0.99077) (3,0.99055) (4,0.99149) (5,0.9902) (5,0.98507) (4,0.9877) (3,0.98285) (2,0.98366) (1,0.98458) (0,0.98736)} -- cycle;
\addplot[fill=black!55, fill opacity=0.22, draw=none, no markers, forget plot] coordinates {(0,0.96804) (1,0.95513) (2,0.9539) (3,0.95558) (4,0.95772) (5,0.96251) (5,0.95247) (4,0.9437) (3,0.93025) (2,0.92556) (1,0.9289) (0,0.93243)} -- cycle;
\addplot[fill=black!55, fill opacity=0.22, draw=none, no markers, forget plot] coordinates {(0,0.96536) (1,0.96495) (2,0.96386) (3,0.965) (4,0.96602) (5,0.96507) (5,0.95971) (4,0.96013) (3,0.95532) (2,0.95502) (1,0.95557) (0,0.95438)} -- cycle;
\addplot[fill=parents, fill opacity=0.22, draw=none, no markers, forget plot] coordinates {(0,0.99668) (1,0.99243) (2,0.99066) (3,0.98873) (4,0.99141) (5,0.99273) (5,0.98856) (4,0.98759) (3,0.97868) (2,0.98239) (1,0.98516) (0,0.9917)} -- cycle;
\addplot[mPCMCI] coordinates {(0,1.000) (1,0.562) (2,0.416) (3,0.304) (4,0.252) (5,0.205)}; \addlegendentry{PCMCI}
\addplot[glaucous, thin, mark=triangle, mark size=3.0pt, mark options={solid}] coordinates {(0,0.973) (1,0.681) (2,0.549) (3,0.388) (4,0.282) (5,0.206)}; \addlegendentry{CD-NOD}
\addplot[mDYNO] coordinates {(0,1.000) (1,0.808) (2,0.685) (3,0.486) (4,0.330) (5,0.148)}; \addlegendentry{DYNOTEARS}
\addplot[mFullCI] coordinates {(0,0.997) (1,0.997) (2,0.998) (3,0.997) (4,0.997) (5,0.998)}; \addlegendentry{FullCI}
\addplot[mIAMB] coordinates {(0,0.997) (1,0.997) (2,0.998) (3,0.997) (4,0.997) (5,0.997)}; \addlegendentry{IAMB}
\addplot[mIPaDL] coordinates {(0,0.992) (1,0.989) (2,0.987) (3,0.987) (4,0.990) (5,0.988)}; \addlegendentry{IPaDL}
\addplot[mDFPC] coordinates {(0,0.953) (1,0.942) (2,0.940) (3,0.944) (4,0.951) (5,0.958)}; \addlegendentry{DF-PC}
\addplot[mLasso] coordinates {(0,0.960) (1,0.960) (2,0.960) (3,0.961) (4,0.963) (5,0.962)}; \addlegendentry{Lasso}
\addplot[mITPD] coordinates {(0,0.994) (1,0.989) (2,0.987) (3,0.984) (4,0.989) (5,0.991)};
\end{axis}
\end{tikzpicture}
    \caption{Mean F1 against the share of edges redrawn at each time step with $d = 2$, $M = 2000$. Share 0 is the stationary graph. Each point is a mean over 20 graphs, and the shaded bands are 95\% intervals.}
    \label{fig:changing}
\end{figure}
\begin{figure}[h] 
    \centering
\begin{tikzpicture}[scale=0.9]
\begin{axis}[
    name=frA, 
    resaxis, width=0.47\linewidth, height=5.2cm,
    xmin=-0.3, xmax=4.3, ymin=0, ymax=1.06,
    xtick={0,1,2,3,4}, xticklabels={50,100,200,500,2000}, ytick={0,0.2,0.4,0.6,0.8,1},
    xlabel={Observations $M$}, ylabel={F1 on all targets},
    title={$N = 20$}, legend columns=4, legend style={at={(1.08,-0.3)}, anchor=north},
]
\addplot[fill=glaucous, fill opacity=0.22, draw=none, no markers, forget plot] coordinates {(0,0.12578) (1,0.16834) (2,0.19593) (3,0.20746) (4,0.20783) (4,0.19957) (3,0.19935) (2,0.18711) (1,0.15534) (0,0.10394)} -- cycle;
\addplot[fill=glaucous, fill opacity=0.22, draw=none, no markers, forget plot] coordinates {(0,0.10099) (1,0.078363) (2,0.068096) (3,0.054086) (4,0.044713) (4,0.029643) (3,0.036324) (2,0.045407) (1,0.059273) (0,0.077927)} -- cycle;
\addplot[fill=black!55, fill opacity=0.22, draw=none, no markers, forget plot] coordinates {(0,0.077749) (1,0.23833) (2,0.48942) (3,0.94115) (4,0.99843) (4,0.99745) (3,0.93453) (2,0.47474) (1,0.21624) (0,0.064317)} -- cycle;
\addplot[fill=black!55, fill opacity=0.22, draw=none, no markers, forget plot] coordinates {(0,0.44599) (1,0.75596) (2,0.94247) (3,0.99629) (4,0.99846) (4,0.99733) (3,0.9936) (2,0.93803) (1,0.73573) (0,0.43025)} -- cycle;
\addplot[fill=black!55, fill opacity=0.22, draw=none, no markers, forget plot] coordinates {(0,0.49958) (1,0.75761) (2,0.87532) (3,0.97085) (4,0.99804) (4,0.99613) (3,0.96695) (2,0.86505) (1,0.73984) (0,0.47033)} -- cycle;
\addplot[fill=black!55, fill opacity=0.22, draw=none, no markers, forget plot] coordinates {(0,0.41816) (1,0.64755) (2,0.80627) (3,0.93128) (4,0.98872) (4,0.98441) (3,0.92219) (2,0.79553) (1,0.63477) (0,0.39902)} -- cycle;
\addplot[fill=black!55, fill opacity=0.22, draw=none, no markers, forget plot] coordinates {(0,0.52838) (1,0.78844) (2,0.91392) (3,0.96523) (4,0.97396) (4,0.96893) (3,0.95886) (2,0.90261) (1,0.77196) (0,0.50185)} -- cycle;
\addplot[fill=parents, fill opacity=0.22, draw=none, no markers, forget plot] coordinates {(0,0.4264) (1,0.78312) (2,0.94344) (3,0.99318) (4,0.99861) (4,0.99563) (3,0.99021) (2,0.93692) (1,0.763) (0,0.39462)} -- cycle;
\addplot[mPCMCI] coordinates {(0,0.115) (1,0.162) (2,0.192) (3,0.204) (4,0.204)}; \addlegendentry{PCMCI}
\addplot[mDYNO] coordinates {(0,0.090) (1,0.068) (2,0.057) (3,0.045) (4,0.037)}; \addlegendentry{DYNOTEARS}
\addplot[mFullCI] coordinates {(0,0.072) (1,0.227) (2,0.483) (3,0.938) (4,0.998)}; \addlegendentry{FullCI}
\addplot[mIAMB] coordinates {(0,0.437) (1,0.747) (2,0.940) (3,0.995) (4,0.998)}; \addlegendentry{IAMB}
\addplot[mIPaDL] coordinates {(0,0.484) (1,0.749) (2,0.870) (3,0.969) (4,0.997)}; \addlegendentry{IPaDL}
\addplot[mDFPC] coordinates {(0,0.409) (1,0.641) (2,0.801) (3,0.927) (4,0.986)}; \addlegendentry{DF-PC}
\addplot[mLasso] coordinates {(0,0.515) (1,0.781) (2,0.908) (3,0.962) (4,0.971)}; \addlegendentry{Lasso}
\addplot[mITPD] coordinates {(0,0.411) (1,0.774) (2,0.940) (3,0.992) (4,0.997)}; \addlegendentry{ITPD}
\end{axis}
\begin{axis}[
     at={(frA.south east)}, anchor=south west, xshift=1.0cm,
    resaxis, width=0.47\linewidth, height=5.2cm,
    xmin=-0.3, xmax=4.3, ymin=0, ymax=1.06,
    xtick={0,1,2,3,4}, xticklabels={50,100,200,500,2000}, ytick={0,0.2,0.4,0.6,0.8,1},
    xlabel={Observations $M$}, 
    title={$N = 80$}, 
]
\addplot[fill=glaucous, fill opacity=0.22, draw=none, no markers, forget plot] coordinates {(0,0.054708) (1,0.081394) (2,0.10365) (3,0.11734) (3,0.11339) (2,0.10074) (1,0.076978) (0,0.049446)} -- cycle;
\addplot[fill=glaucous, fill opacity=0.22, draw=none, no markers, forget plot] coordinates {(0,0.042363) (1,0.020941) (2,0.0090427) (3,0.0030994) (4,0.0029245) (4,0.0002603) (3,0.001252) (2,0.0058071) (1,0.016) (0,0.036442)} -- cycle;
\addplot[fill=black!55, fill opacity=0.22, draw=none, no markers, forget plot] coordinates {(0,0) (1,0.043078) (2,0.16024) (3,0.40166) (4,0.9957) (4,0.99456) (3,0.39097) (2,0.15429) (1,0.0372) (0,0)} -- cycle;
\addplot[fill=black!55, fill opacity=0.22, draw=none, no markers, forget plot] coordinates {(0,0.28729) (1,0.67488) (2,0.92378) (3,0.99348) (4,0.99803) (4,0.99714) (3,0.99256) (2,0.91835) (1,0.66276) (0,0.27287)} -- cycle;
\addplot[fill=black!55, fill opacity=0.22, draw=none, no markers, forget plot] coordinates {(0,0.34774) (1,0.64307) (2,0.8046) (3,0.96582) (4,0.99899) (4,0.99832) (3,0.96277) (2,0.7969) (1,0.6341) (0,0.33891)} -- cycle;
\addplot[fill=black!55, fill opacity=0.22, draw=none, no markers, forget plot] coordinates {(0,0.29623) (1,0.49479) (2,0.75343) (3,0.9387) (4,0.99502) (4,0.99409) (3,0.93569) (2,0.74765) (1,0.48551) (0,0.28791)} -- cycle;
\addplot[fill=black!55, fill opacity=0.22, draw=none, no markers, forget plot] coordinates {(0,0.39105) (1,0.71569) (2,0.89956) (3,0.97758) (4,0.98745) (4,0.98505) (3,0.97341) (2,0.89325) (1,0.70526) (0,0.38032)} -- cycle;
\addplot[fill=parents, fill opacity=0.22, draw=none, no markers, forget plot] coordinates {(0,0.30743) (1,0.69317) (2,0.91165) (3,0.98982) (4,0.99814) (4,0.99728) (3,0.98831) (2,0.90725) (1,0.68205) (0,0.2932)} -- cycle;
\addplot[mPCMCI] coordinates {(0,0.052) (1,0.079) (2,0.102) (3,0.115)};
\addplot[mDYNO] coordinates {(0,0.039) (1,0.019) (2,0.007) (3,0.002) (4,0.002)};
\addplot[mFullCI] coordinates {(0,0.000) (1,0.040) (2,0.158) (3,0.396) (4,0.995)};
\addplot[mIAMB] coordinates {(0,0.280) (1,0.668) (2,0.921) (3,0.993) (4,0.998)};
\addplot[mIPaDL] coordinates {(0,0.343) (1,0.638) (2,0.801) (3,0.964) (4,0.999)};
\addplot[mDFPC] coordinates {(0,0.292) (1,0.490) (2,0.750) (3,0.937) (4,0.995)};
\addplot[mLasso] coordinates {(0,0.385) (1,0.710) (2,0.897) (3,0.976) (4,0.986)};
\addplot[mITPD] coordinates {(0,0.300) (1,0.687) (2,0.909) (3,0.989) (4,0.998)};
\end{axis}
\end{tikzpicture}

\caption{Mean F1 on all targets against the number of observations $M$ for $N = 20$ (left) and $N = 80$ (right) (nonstationary graph, linear-Gaussian data, $T = 16$, lag 1, $d = 2$; 10 graphs per point). 
A target at the last step has 299 candidates at $N = 20$ and 1{,}199 at $N = 80$. PCMCI did not finish at $N = 80$, $M = 2000$. Each point is a mean over 10 graphs, and the shaded bands are 95\% intervals.}
    \label{fig:samples}
\end{figure}
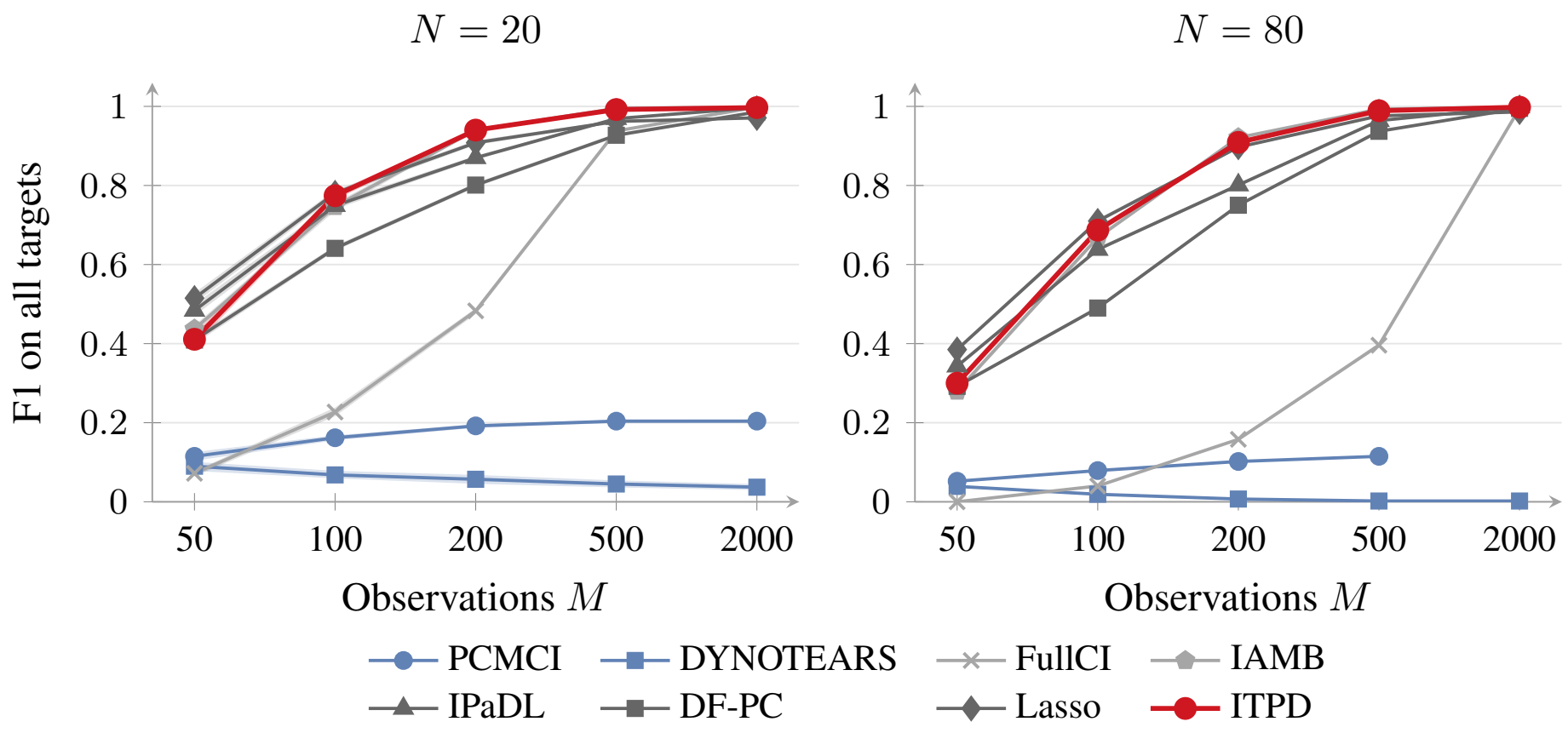

\clearpage

\begin{table}[h]
    \centering
    \footnotesize
    \begin{tabular}{l c c c c c}
    \toprule
    $M$ & Settings & F1 with $\alpha / m$ & F1 with extended BIC & Change & Higher / level / lower \\
    \midrule
    50 & 1 & 0.300 & 0.436 & $+0.136$ & 1 / 0 / 0 \\
    200 & 7 & 0.684 & 0.718 & $+0.034$ & 3 / 1 / 3 \\
    500 & 12 & 0.886 & 0.875 & $-0.012$ & 2 / 0 / 10 \\
    $\geq 2000$ & 14 & 0.922 & 0.910 & $-0.012$ & 0 / 0 / 14 \\
    \bottomrule
    \end{tabular}
    \caption{Mean F1 of ITPD with $\alpha_2 = \alpha / m$ and with the extended-BIC level $P(\chi^2_1 > \log(M m))$ in Phase II, on 34 held-out settings grouped by $M$ (3 or 5 graphs per setting). The last column counts the settings where the extended-BIC level was higher, level, or lower by a paired 95\% interval.}
    \label{tab:extended_bic}
\end{table}

\begin{table}[h]
    \centering
    \footnotesize
    \setlength{\tabcolsep}{2.8pt}
    \begin{tabular}{l cccc c cccc}
    \toprule
    & \multicolumn{4}{c}{F1, $M = 500$} && \multicolumn{4}{c}{F1, $M = 2000$} \\
    \cmidrule(lr){2-5} \cmidrule(lr){7-10}
    Method & $d = 1$ & $d = 2$ & $d = 3$ & $d = 5$ && $d = 1$ & $d = 2$ & $d = 3$ & $d = 5$ \\
    \midrule
    ITPD & \textbf{0.995}$_{\pm .002}$ & \underline{0.992}$_{\pm .002}$ & \underline{0.985}$_{\pm .001}$ & \underline{0.961}$_{\pm .002}$ && \underline{0.997}$_{\pm .001}$ & \underline{0.997}$_{\pm .001}$ & \underline{0.996}$_{\pm .002}$ & 0.991$_{\pm .002}$ \\
    IAMB & \underline{0.994}$_{\pm .002}$ & \textbf{0.994}$_{\pm .002}$ & \textbf{0.991}$_{\pm .001}$ & \textbf{0.972}$_{\pm .002}$ && 0.996$_{\pm .002}$ & \textbf{0.998}$_{\pm .001}$ & \textbf{0.998}$_{\pm .001}$ & \textbf{0.999}$_{\pm .001}$ \\
    IPaDL & 0.988$_{\pm .003}$ & 0.967$_{\pm .003}$ & 0.936$_{\pm .004}$ & 0.913$_{\pm .003}$ && \textbf{0.998}$_{\pm .001}$ & \underline{0.997}$_{\pm .001}$ & 0.993$_{\pm .002}$ & 0.982$_{\pm .002}$ \\
    FullCI & 0.967$_{\pm .004}$ & 0.935$_{\pm .004}$ & 0.894$_{\pm .005}$ & 0.815$_{\pm .004}$ && 0.996$_{\pm .002}$ & \textbf{0.998}$_{\pm .001}$ & \textbf{0.998}$_{\pm .001}$ & \underline{0.998}$_{\pm .001}$ \\
    DF-PC & 0.967$_{\pm .004}$ & 0.926$_{\pm .004}$ & 0.895$_{\pm .005}$ & 0.837$_{\pm .003}$ && 0.996$_{\pm .002}$ & 0.985$_{\pm .002}$ & 0.966$_{\pm .002}$ & 0.931$_{\pm .004}$ \\
    Lasso & 0.937$_{\pm .005}$ & 0.963$_{\pm .003}$ & 0.955$_{\pm .003}$ & 0.916$_{\pm .003}$ && 0.940$_{\pm .005}$ & 0.970$_{\pm .002}$ & 0.971$_{\pm .002}$ & 0.961$_{\pm .002}$ \\
    PCMCI & 0.158$_{\pm .005}$ & 0.205$_{\pm .004}$ & 0.264$_{\pm .005}$ & 0.364$_{\pm .005}$ && 0.151$_{\pm .004}$ & 0.207$_{\pm .004}$ & 0.273$_{\pm .004}$ & 0.396$_{\pm .004}$ \\
    CD-NOD & 0.160$_{\pm .004}$ & 0.205$_{\pm .005}$ & 0.253$_{\pm .006}$ & 0.337$_{\pm .005}$ && 0.152$_{\pm .003}$ & 0.207$_{\pm .004}$ & 0.269$_{\pm .005}$ & 0.377$_{\pm .005}$ \\
    DYNOTEARS & 0.029$_{\pm .009}$ & 0.045$_{\pm .008}$ & 0.050$_{\pm .007}$ & 0.062$_{\pm .007}$ && 0.029$_{\pm .009}$ & 0.038$_{\pm .008}$ & 0.045$_{\pm .007}$ & 0.055$_{\pm .007}$ \\
    \bottomrule
    \end{tabular}
    \caption{Mean F1 in the main finite data setting. Each cell is a mean over 20 graphs; subscripts are the half-widths of the 95\% intervals.}
    \label{tab:all}
\end{table}
\begin{table}[!htb]
    \centering
    \footnotesize
    \setlength{\tabcolsep}{2.5pt}
    \begin{tabular}{l cccc c cccc c cccc}
    \toprule
    & \multicolumn{4}{c}{Unique CI tests per candidate} && \multicolumn{4}{c}{Mean conditioning set} && \multicolumn{4}{c}{Max conditioning set} \\
    \cmidrule(lr){2-5} \cmidrule(lr){7-10} \cmidrule(lr){12-15}
    Method & $d = 1$ & 2 & 3 & 5 && $d = 1$ & 2 & 3 & 5 && $d = 1$ & 2 & 3 & 5 \\
    \midrule
    ITPD & 1.35$_{\pm .01}$ & \textbf{1.38}$_{\pm .01}$ & \textbf{1.40}$_{\pm .01}$ & \textbf{1.42}$_{\pm .01}$ && 6.7$_{\pm .1}$ & 8.2$_{\pm .1}$ & 9.4$_{\pm .1}$ & 11.6$_{\pm .1}$ && 30$_{\pm 1}$ & 33$_{\pm 1}$ & 35$_{\pm 1}$ & 37$_{\pm 2}$ \\
    IAMB & 2.00$_{\pm .03}$ & 2.97$_{\pm .04}$ & 3.97$_{\pm .05}$ & 5.96$_{\pm .06}$ && 1.7$_{\pm .1}$ & \underline{2.3}$_{\pm .1}$ & \underline{2.8}$_{\pm .1}$ & \underline{3.8}$_{\pm .1}$ && \underline{6}$_{\pm 1}$ & \underline{7}$_{\pm 1}$ & \underline{9}$_{\pm 1}$ & \underline{12}$_{\pm 1}$ \\
    IPaDL & \textbf{1.26}$_{\pm .01}$ & \underline{1.39}$_{\pm .01}$ & \underline{1.48}$_{\pm .01}$ & \underline{1.56}$_{\pm .01}$ && \underline{1.3}$_{\pm .1}$ & 3.3$_{\pm .2}$ & 5.2$_{\pm .2}$ & 7.2$_{\pm .2}$ && 42$_{\pm 3}$ & 55$_{\pm 2}$ & 64$_{\pm 3}$ & 71$_{\pm 2}$ \\
    FullCI & 1.00$_{\pm 0}$ & 1.00$_{\pm 0}$ & 1.00$_{\pm 0}$ & 1.00$_{\pm 0}$ && 206$_{\pm 0}$ & 206$_{\pm 0}$ & 206$_{\pm 0}$ & 206$_{\pm 0}$ && 299$_{\pm 0}$ & 299$_{\pm 0}$ & 299$_{\pm 0}$ & 299$_{\pm 0}$ \\
    DF-PC & \underline{1.31}$_{\pm .02}$ & 1.83$_{\pm .03}$ & 2.58$_{\pm .05}$ & 14.61$_{\pm .43}$ && \textbf{0.3}$_{\pm .1}$ & \textbf{0.6}$_{\pm .1}$ & \textbf{0.9}$_{\pm .1}$ & \textbf{1.8}$_{\pm .1}$ && \textbf{5}$_{\pm 1}$ & \textbf{6}$_{\pm 1}$ & \textbf{7}$_{\pm 1}$ & \textbf{9}$_{\pm 1}$ \\
    PCMCI & -- & -- & -- & -- && 5.6$_{\pm .2}$ & 6.4$_{\pm .2}$ & 7.8$_{\pm .1}$ & 8.1$_{\pm .2}$ && 21$_{\pm 1}$ & 25$_{\pm 1}$ & 32$_{\pm 1}$ & 33$_{\pm 1}$ \\
    \bottomrule
    \end{tabular}
    \caption{Unique CI tests per candidate and conditioning set sizes in the main finite data setting with $M = 2000$. Each cell is a mean over 20 graphs; subscripts are the half-widths of the 95\% intervals.}
    \label{tab:tests}
\end{table}
\begin{table}[!htb]
    \centering
    \footnotesize
    \setlength{\tabcolsep}{2.8pt}
    \begin{tabular}{l cccc c cccc}
    \toprule
    & \multicolumn{4}{c}{Stationary graph} && \multicolumn{4}{c}{Nonstationary graph} \\
    \cmidrule(lr){2-5} \cmidrule(lr){7-10}
    $N$, $T$ & 10, 8 & 10, 8 & 10, 16 & 20, 8 && 10, 8 & 10, 8 & 20, 8 & 20, 8 \\
    $M$ & 500 & 2000 & 2000 & 2000 && 500 & 2000 & 500 & 2000 \\
    \midrule
    ITPD & 0.963$_{\pm .009}$ & \textbf{0.991}$_{\pm .005}$ & \textbf{0.985}$_{\pm .007}$ & 0.984$_{\pm .009}$ && \underline{0.957}$_{\pm .008}$ & \underline{0.989}$_{\pm .006}$ & \underline{0.957}$_{\pm .006}$ & \textbf{0.989}$_{\pm .003}$ \\
    IAMB & 0.974$_{\pm .009}$ & 0.981$_{\pm .007}$ & 0.973$_{\pm .008}$ & 0.975$_{\pm .007}$ && \textbf{0.967}$_{\pm .005}$ & 0.984$_{\pm .005}$ & \textbf{0.963}$_{\pm .003}$ & \underline{0.977}$_{\pm .004}$ \\
    IPaDL & 0.932$_{\pm .022}$ & 0.988$_{\pm .009}$ & 0.977$_{\pm .012}$ & \textbf{0.990}$_{\pm .006}$ && 0.923$_{\pm .014}$ & 0.981$_{\pm .012}$ & 0.936$_{\pm .012}$ & \textbf{0.989}$_{\pm .002}$ \\
    FullCI & 0.944$_{\pm .014}$ & \underline{0.990}$_{\pm .005}$ & -- & -- && 0.937$_{\pm .006}$ & \textbf{0.990}$_{\pm .003}$ & 0.938$_{\pm .008}$ & -- \\
    DF-PC & 0.897$_{\pm .027}$ & 0.921$_{\pm .025}$ & 0.902$_{\pm .022}$ & 0.929$_{\pm .015}$ && 0.889$_{\pm .013}$ & -- & 0.896$_{\pm .011}$ & -- \\
    Lasso & 0.898$_{\pm .012}$ & 0.861$_{\pm .017}$ & 0.834$_{\pm .019}$ & 0.876$_{\pm .025}$ && 0.903$_{\pm .009}$ & 0.880$_{\pm .013}$ & 0.921$_{\pm .009}$ & 0.902$_{\pm .008}$ \\
    PCMCI & \underline{0.986}$_{\pm .015}$ & 0.975$_{\pm .017}$ & 0.966$_{\pm .023}$ & 0.947$_{\pm .014}$ && 0.363$_{\pm .020}$ & 0.370$_{\pm .020}$ & -- & -- \\
    DYNOTEARS & \textbf{0.989}$_{\pm .009}$ & \textbf{0.991}$_{\pm .008}$ & \underline{0.978}$_{\pm .016}$ & \underline{0.986}$_{\pm .012}$ && 0.324$_{\pm .023}$ & 0.318$_{\pm .026}$ & 0.243$_{\pm .015}$ & 0.242$_{\pm .013}$ \\
    \bottomrule
    \end{tabular}
    \caption{Mean F1 with nonlinear causal mechanisms at $d = 2$, lag 1. A dash indicates that the runs did not finish within the given time budget (4 hours). Each cell is a mean over 10 graphs; subscripts are the half-widths of the 95\% intervals. 
    } 
    \label{tab:nonlinear}
\end{table}
\begin{table}[t]
    \centering
    \footnotesize
    \setlength{\tabcolsep}{4.5pt}
    \begin{tabular}{l cccc c cccc}
    \toprule
    & \multicolumn{4}{c}{F1 on common targets} && \multicolumn{4}{c}{Unique CI tests per candidate} \\
    \cmidrule(lr){2-5} \cmidrule(lr){7-10}
    & \multicolumn{2}{c}{$N = 50$} & \multicolumn{2}{c}{$N = 100$} && \multicolumn{2}{c}{$N = 50$} & \multicolumn{2}{c}{$N = 100$} \\
    Method & $M = 500$ & 2000 & 500 & 2000 && $M = 500$ & 2000 & 500 & 2000 \\
    \midrule
    ITPD & \underline{0.994}$_{\pm .001}$ & \textbf{0.998}$_{\pm .001}$ & \underline{0.992}$_{\pm .003}$ & \underline{0.998}$_{\pm .001}$ && 1.15$_{\pm .01}$ & \underline{1.18}$_{\pm .01}$ & 1.08$_{\pm .01}$ & \underline{1.10}$_{\pm .01}$ \\
    IAMB & \textbf{0.995}$_{\pm .001}$ & \textbf{0.998}$_{\pm .001}$ & \textbf{0.994}$_{\pm .002}$ & \underline{0.998}$_{\pm .001}$ && 2.99$_{\pm .04}$ & 3.00$_{\pm .05}$ & 2.98$_{\pm .04}$ & 3.00$_{\pm .03}$ \\
    IPaDL & 0.971$_{\pm .003}$ & \textbf{0.998}$_{\pm .001}$ & 0.973$_{\pm .005}$ & \textbf{0.999}$_{\pm .001}$ && \underline{1.09}$_{\pm .01}$ & \textbf{1.16}$_{\pm .01}$ & \underline{1.07}$_{\pm .01}$ & \underline{1.10}$_{\pm .01}$ \\
    FullCI & 0.804$_{\pm .009}$ & \underline{0.997}$_{\pm .001}$ & 0.836$_{\pm .011}$ & 0.955$_{\pm .004}$ && 0.24$_{\pm 0}$ & 1.00$_{\pm 0}$ & 0.05$_{\pm 0}$ & 1.00$_{\pm 0}$ \\
    DF-PC & 0.940$_{\pm .004}$ & 0.993$_{\pm .001}$ & 0.949$_{\pm .004}$ & 0.996$_{\pm .001}$ && \textbf{1.04}$_{\pm .01}$ & 1.20$_{\pm .02}$ & \textbf{1.02}$_{\pm .01}$ & \textbf{1.08}$_{\pm .01}$ \\
    Lasso & 0.972$_{\pm .005}$ & 0.983$_{\pm .004}$ & 0.979$_{\pm .003}$ & 0.988$_{\pm .001}$ && -- & -- & -- & -- \\
    PCMCI & 0.115$_{\pm .005}$ & 0.125$_{\pm .003}$ & 0.080$_{\pm .002}$ & 0.107$_{\pm .001}$ && -- & -- & -- & -- \\
    CD-NOD & 0.119$_{\pm .006}$ & 0.125$_{\pm .003}$ & 0.089$_{\pm .005}$ & 0.106$_{\pm .001}$ && -- & -- & -- & -- \\
    \bottomrule
    \end{tabular}
    \caption{Mean F1 and unique CI tests per candidate on larger graphs. At $M = 500$, the common targets are at early time steps. FullCI tests only the targets with fewer candidates than observations. Each cell is a mean over 5 graphs; subscripts are the half-widths of the 95\% intervals.}
    \label{tab:scale}
\end{table}
\begin{table}[!htb]
    \centering
    \footnotesize
    \begin{tabular}{l ccc c ccc}
    \toprule
    & \multicolumn{3}{c}{$d = 2$} && \multicolumn{3}{c}{$d = 5$} \\
    \cmidrule(lr){2-4} \cmidrule(lr){6-8}
    $N$ & ITPD (s) & IAMB (s) & ITPD / IAMB && ITPD (s) & IAMB (s) & ITPD / IAMB \\
    \midrule
    20 & 3.4 to 3.6 & 5.5 & 0.62 to 0.65 && 3.8 to 4.0 & 11 to 12 & 0.34 \\
    50 & 27 to 29 & 56 to 57 & 0.48 to 0.50 && 31 to 33 & 111 to 119 & 0.28 \\
    100 & 96 to 99 & 230 to 231 & 0.41 to 0.43 && 109 to 115 & 450 to 490 & 0.24 \\
    \bottomrule
    \end{tabular}
    \caption{Seconds per run of ITPD and IAMB with plain Fisher-z. Each setting has 5 graphs and 3 timed rounds per graph; the 95\% interval of the ratio is below 1 in every setting.}
    \label{tab:runtime}
\end{table}
\begin{table}[!htb]
    \centering
    \footnotesize
    \setlength{\tabcolsep}{3.5pt}
    \begin{tabular}{l c c cccc}
    \toprule
    & && \multicolumn{4}{c}{No lag-1 self edge} \\
    & && \multicolumn{4}{c}{(lag-1 edges)} \\
    \cmidrule(lr){4-7}
    Share & 0 && 0.1 & 0.25 & 0.5 & 1 \\
    \midrule
    ITPD & \underline{0.997}$_{\pm .001}$ && \textbf{0.998}$_{\pm .001}$ & \textbf{0.997}$_{\pm .001}$ & \textbf{0.997}$_{\pm .001}$ & \underline{0.997}$_{\pm .001}$ \\
    IAMB & \textbf{0.998}$_{\pm .001}$ && \underline{0.997}$_{\pm .001}$ & \textbf{0.997}$_{\pm .001}$ & \textbf{0.997}$_{\pm .001}$ & \underline{0.997}$_{\pm .001}$ \\
    FullCI & \textbf{0.998}$_{\pm .001}$ && \underline{0.997}$_{\pm .001}$ & \textbf{0.997}$_{\pm .001}$ & \textbf{0.997}$_{\pm .001}$ & \textbf{0.998}$_{\pm .001}$ \\
    Lasso & 0.970$_{\pm .002}$ && 0.972$_{\pm .003}$ & \underline{0.972}$_{\pm .003}$ & \underline{0.973}$_{\pm .003}$ & 0.979$_{\pm .002}$ \\
    PCMCI & 0.207$_{\pm .004}$ && 0.207$_{\pm .004}$ & 0.207$_{\pm .005}$ & 0.209$_{\pm .004}$ & 0.213$_{\pm .004}$ \\
    \bottomrule
    \end{tabular}
    \caption{Mean F1 with one assumption violated with $d = 2$, $M = 2000$. $N = 20$, $T = 16$. Each cell is a mean over 20 graphs; subscripts are the half-widths of the 95\% intervals. 
    }
    \label{tab:violations}
\end{table}

\begin{table}[!htb]
    \centering
    \footnotesize
    \setlength{\tabcolsep}{3.5pt}
    \begin{tabular}{l ccc c ccc}
    \toprule
    & \multicolumn{3}{c}{Edges within a time step} && \multicolumn{3}{c}{Edges within a time step} \\
    & \multicolumn{3}{c}{(lag-1 edges)} && \multicolumn{3}{c}{(Markov boundary)} \\
    \cmidrule(lr){2-4} \cmidrule(lr){6-8}
    Share & 0.1 & 0.25 & 0.5 && 0.1 & 0.25 & 0.5 \\
    \midrule
    ITPD & 0.889$_{\pm .003}$ & 0.744$_{\pm .003}$ & 0.527$_{\pm .004}$ && \underline{0.968}$_{\pm .004}$ & \underline{0.894}$_{\pm .006}$ & 0.733$_{\pm .011}$ \\
    IAMB & 0.883$_{\pm .004}$ & 0.732$_{\pm .003}$ & 0.511$_{\pm .004}$ && \textbf{0.979}$_{\pm .003}$ & \textbf{0.917}$_{\pm .005}$ & \textbf{0.763}$_{\pm .011}$ \\
    FullCI & 0.901$_{\pm .004}$ & 0.774$_{\pm .004}$ & 0.574$_{\pm .005}$ && 0.961$_{\pm .004}$ & 0.874$_{\pm .006}$ & 0.690$_{\pm .011}$ \\
    Lasso & 0.849$_{\pm .004}$ & 0.687$_{\pm .004}$ & 0.452$_{\pm .005}$ && 0.945$_{\pm .004}$ & 0.885$_{\pm .005}$ & \underline{0.754}$_{\pm .009}$ \\
    PCMCI & 0.189$_{\pm .004}$ & 0.163$_{\pm .004}$ & 0.114$_{\pm .003}$ && 0.227$_{\pm .004}$ & 0.265$_{\pm .005}$ & 0.342$_{\pm .009}$ \\
    \bottomrule
    \end{tabular}
    \caption{Same setting as Table~\ref{tab:violations}, with edges moved into the same time step, scored on the remaining lag-1 edges (left, unranked) and on the Markov boundary among earlier variables (right). Each cell is a mean over 20 graphs; subscripts are the half-widths of the 95\% intervals.}
    \label{tab:violations_contemp}
\end{table}
\begin{table}[!htb]
    \centering
    \footnotesize
    \setlength{\tabcolsep}{3pt}
    \begin{tabular}{l ccc c ccc}
    \toprule
    & \multicolumn{7}{c}{Regulatory edges} \\
    \cmidrule(lr){2-8}
    & \multicolumn{3}{c}{Network 3} && \multicolumn{3}{c}{Network 5} \\
    $M$ & 200 & 500 & 2000 && 200 & 500 & 2000 \\
    \midrule
    ITPD & 0.518$_{\pm .022}$ & \textbf{0.683}$_{\pm .006}$ & 0.654$_{\pm .003}$ && 0.464$_{\pm .013}$ & \textbf{0.541}$_{\pm .007}$ & 0.483$_{\pm .004}$ \\
    IAMB & 0.510$_{\pm .024}$ & 0.644$_{\pm .006}$ & 0.639$_{\pm .003}$ && 0.423$_{\pm .021}$ & 0.500$_{\pm .008}$ & 0.468$_{\pm .004}$ \\
    IPaDL & 0.439$_{\pm .013}$ & 0.593$_{\pm .008}$ & 0.652$_{\pm .004}$ && 0.352$_{\pm .013}$ & 0.472$_{\pm .010}$ & \underline{0.542}$_{\pm .006}$ \\
    FullCI & 0.179$_{\pm .014}$ & 0.346$_{\pm .007}$ & 0.629$_{\pm .006}$ && 0.159$_{\pm .020}$ & 0.315$_{\pm .009}$ & 0.530$_{\pm .006}$ \\
    DF-PC & 0.530$_{\pm .028}$ & 0.579$_{\pm .006}$ & \textbf{0.747}$_{\pm .006}$ && 0.406$_{\pm .014}$ & \underline{0.536}$_{\pm .009}$ & \textbf{0.599}$_{\pm .008}$ \\
    Lasso & 0.431$_{\pm .017}$ & 0.494$_{\pm .006}$ & 0.451$_{\pm .004}$ && 0.363$_{\pm .018}$ & 0.400$_{\pm .009}$ & 0.343$_{\pm .004}$ \\
    PCMCI & 0.609$_{\pm .011}$ & 0.539$_{\pm .016}$ & 0.454$_{\pm .008}$ && 0.504$_{\pm .016}$ & 0.436$_{\pm .013}$ & 0.356$_{\pm .005}$ \\
    CD-NOD & \underline{0.681}$_{\pm .015}$ & 0.608$_{\pm .012}$ & 0.512$_{\pm .009}$ && \textbf{0.527}$_{\pm .014}$ & 0.452$_{\pm .012}$ & 0.375$_{\pm .015}$ \\
    DYNOTEARS & \textbf{0.689}$_{\pm .024}$ & \underline{0.654}$_{\pm .025}$ & \underline{0.660}$_{\pm .018}$ && \underline{0.526}$_{\pm 0}$ & 0.526$_{\pm 0}$ & 0.526$_{\pm 0}$ \\
    \bottomrule
    \end{tabular}
    \caption{Mean F1 on re-simulated DREAM4 gene regulatory networks. In Table~\ref{tab:gnw_indirect}, indirect pairs counted neither as true nor as false. Each cell is a mean over 10 data sets; subscripts are the half-widths of the 95\% intervals. 
}
    \label{tab:gnw}
\end{table}

\begin{table}[!htb]
    \centering
    \footnotesize
    \setlength{\tabcolsep}{3pt}
    \begin{tabular}{l ccc c ccc}
    \toprule
    & \multicolumn{7}{c}{Regulatory edges, indirect pairs left out} \\
    \cmidrule(lr){2-8}
    & \multicolumn{3}{c}{Network 3} && \multicolumn{3}{c}{Network 5} \\
    $M$ & 200 & 500 & 2000 && 200 & 500 & 2000 \\
    \midrule
    ITPD & 0.539$_{\pm .024}$ & 0.731$_{\pm .005}$ & \textbf{0.849}$_{\pm .002}$ && 0.515$_{\pm .011}$ & \underline{0.676}$_{\pm .007}$ & 0.731$_{\pm .005}$ \\
    IAMB & 0.521$_{\pm .025}$ & 0.692$_{\pm .006}$ & \underline{0.832}$_{\pm .003}$ && 0.478$_{\pm .021}$ & 0.639$_{\pm .008}$ & 0.703$_{\pm .008}$ \\
    IPaDL & 0.464$_{\pm .015}$ & 0.611$_{\pm .009}$ & 0.765$_{\pm .003}$ && 0.368$_{\pm .013}$ & 0.541$_{\pm .010}$ & \underline{0.743}$_{\pm .006}$ \\
    FullCI & 0.182$_{\pm .015}$ & 0.351$_{\pm .008}$ & 0.719$_{\pm .006}$ && 0.164$_{\pm .021}$ & 0.332$_{\pm .010}$ & 0.704$_{\pm .006}$ \\
    DF-PC & 0.540$_{\pm .028}$ & 0.586$_{\pm .007}$ & 0.790$_{\pm .007}$ && 0.421$_{\pm .012}$ & 0.583$_{\pm .007}$ & \textbf{0.754}$_{\pm .007}$ \\
    Lasso & 0.490$_{\pm .023}$ & 0.597$_{\pm .007}$ & 0.603$_{\pm .006}$ && 0.476$_{\pm .024}$ & 0.572$_{\pm .015}$ & 0.528$_{\pm .007}$ \\
    PCMCI & \textbf{0.851}$_{\pm .014}$ & \textbf{0.785}$_{\pm .019}$ & 0.682$_{\pm .013}$ && \textbf{0.850}$_{\pm .030}$ & \textbf{0.801}$_{\pm .019}$ & 0.670$_{\pm .016}$ \\
    CD-NOD & \underline{0.823}$_{\pm .017}$ & \underline{0.751}$_{\pm .015}$ & 0.681$_{\pm .014}$ && \underline{0.771}$_{\pm .030}$ & 0.661$_{\pm .020}$ & 0.549$_{\pm .020}$ \\
    DYNOTEARS & 0.689$_{\pm .024}$ & 0.654$_{\pm .025}$ & 0.660$_{\pm .018}$ && 0.588$_{\pm 0}$ & 0.588$_{\pm 0}$ & 0.588$_{\pm 0}$ \\
    \bottomrule
    \end{tabular}
    \caption{Same runs as Table~\ref{tab:gnw}, with indirect pairs counted neither as true nor as false. Each cell is a mean over 10 data sets; subscripts are the half-widths of the 95\% intervals.}
    \label{tab:gnw_indirect}
\end{table}


\clearpage

\subsection{Ablations: Negative Results Post-Mortem}
\label{appendix:ablations}

\paragraph{Step-Wise Ablations on Algorithm \ref{alg:itpd}} To provide insight on the genesis of ITPD and the negative results we encountered along the way, we include Table \ref{tab:ablation}. This shows the effects of ablating specific steps on the cardinality of $\mathbf{S}$, the number of tests performed, and F1 scores. These results were used to justify our design choices on the final draft of Algorithm \ref{alg:itpd}.

\paragraph{Negative Results} Along the way, several additional causal partitioning steps were introduced and phased out due to low yield. In particular, the original \textit{Lines 9 and 10}, which preceded what are now Lines 9 and 10 in Algorithm \ref{alg:itpd}, were defined as follows:
\begin{align*}
    9&': \textbf{if \;} {\cit(Z \not\ind X)} \textbf{\; then \;} & {\color{gray}Z \not\in \z_4.} \\
    10&': \quad \textbf{if \;} \cit(Z \ind Y \mid X) \text{\; \textbf{then} continue \;} & {\color{gray}Z \in \z_5 \not\subset \pa(Y).}
\end{align*}
Line 9' follows from weak transitivity, per Proposition \ref{prop:z4} and Lemma \ref{lemma:z4}.
\begin{proposition}
\label{prop:z4}
    $X \not\ind Y \land Z \ind X \land Z \not\ind Y \Longrightarrow Z \not\ind X\mid Y.$
\end{proposition}
\begin{lemma}[Marginal tests for $\z_4$]
\label{lemma:z4}
    Consider a variable pair $\{X,Y\}$ where $X$ is known to be a direct or indirect cause of $Y$. Let $\z_4 \coloneqq \{Z : Z \ind X \land Z \not\ind X \mid Y\}$. Following from Proposition \ref{prop:z4}, marginal tests $Z \ind X$ and $Z \not\ind Y$ are sufficient to identify $\z_4$.
\end{lemma}
Line 10' is a test for $\z_5$. The theoretical justification for Lines 9' and 10' was the following: (1) Algorithm \ref{alg:itpd} remains sound and complete under their inclusion; (2) $\z_5$ cannot be parents of $Y$ by definition \citep{maasch2024local}, and a first-order CI test to remove them is preferable to a higher-order test at subsequent steps; and (3) in all cases $\z_4 \cap \z_5 = \varnothing$ and it is possible in some cases that $\z_4(X,Y) \cap \pa(Y) \neq \varnothing$, hence gating the first-order test with a zero-order (marginal test) is preferable. In practice, however, these causal partitioning steps provided little added value wrt mean conditioning set size at the expense of extra CI tests (Table~\ref{tab:dropped_steps}). Hence, these were dropped from the final algorithm. 

Additionally, we attempted to define Line 11 as $\widehat{\pa}(Y) \gets \{X\} \cup \{Z : Z \in \mathbf{S} \text{\;if\;} Z \in \mathbf{V}_1\}$ under the premise that any ancestor of $Y$ that is one time step prior must be a parent. In practice, however, this version of Line 11 did not provided added value on finite data (Table~\ref{tab:dropped_steps}). 

\begin{table}[t]
    \centering
    \footnotesize
    \begin{tabular}{l c c c}
    \toprule
    & Algorithm \ref{alg:itpd} & With Lines 9$'$ and 10$'$ & With the Line 11 variant \\
    \midrule
    F1 & 0.813 to 0.997 & 0.813 to 0.997 & 0.610 to 0.881 \\
    Precision & 0.989 to 0.996 & 0.989 to 0.996 & 0.443 to 0.817 \\
    Recall & 0.689 to 1.000 & 0.688 to 1.000 & 0.855 to 1.000 \\
    Unique CI tests per candidate & 1.30 to 1.42 & 1.89 to 2.40 & 1.28 to 1.38 \\
    Mean conditioning set & 5.97 to 11.56 & 3.92 to 6.05 & 6.62 to 11.94 \\
    Max conditioning set & 26.9 to 37.2 & 26.6 to 35.8 & 27.3 to 37.3 \\
    $|\mathbf{S}|$ before Phase II & 12.0 to 18.9 & 11.8 to 17.9 & 12.2 to 18.9 \\
    \bottomrule
    \end{tabular}
    \caption{Steps dropped from Algorithm \ref{alg:itpd}, each added back alone. Each cell is the range of the setting means over the 12 settings of the main setting ($N = 20$, $T = 16$, $d \in \{1, 2, 3, 5\}$, $M \in \{200, 500, 2000\}$; 20 graphs per setting). The Line 11 variant that was run adds the members of $\mathbf{S}$ at the time step of $X$ to $\widehat{\pa}(Y)$ without a test.}
    \label{tab:dropped_steps}
\end{table}

\bigskip

\begin{table}[t]
    \centering
    \scriptsize
    \setlength{\tabcolsep}{3pt}
    \begin{tabular}{@{}l c c c c c c@{}}
    \toprule
    & & \multicolumn{2}{c}{Exact oracle} & \multicolumn{3}{c}{Data} \\
    \cmidrule(lr){3-4} \cmidrule(lr){5-7}
    Change to Algorithm~\ref{alg:itpd} & Exact & $|\mathbf{S}|$ & Tests & F1 & $|\mathbf{S}|$ & Tests \\
    \midrule
    $\z$ in ascending time order & yes & $+63$ to $+137$ & $+47\%$ to $+139\%$ & $-0.004$ to $+0.001$ & $+3$ to $+26$ & $+5\%$ to $+28\%$ \\
    Line 9 given $X$ & yes & $+65$ to $+137$ & $+65\%$ to $+141\%$ & $-0.004$ to $0.000$ & $+6$ to $+29$ & $+7\%$ to $+32\%$ \\
    Line 9 removed & yes & $+85$ to $+137$ & $+20\%$ to $+62\%$ & $-0.003$ to $+0.001$ & $+21$ to $+45$ & $-2\%$ to $+5\%$ \\
    Line 10 removed & yes & $+5$ to $+31$ & $-15\%$ to $-3\%$ & $0.000$ to $+0.012$ & $+11$ to $+14$ & $-9\%$ to $-7\%$ \\
    Lines 8, 9, 10 at level $\alpha$ & yes & -- & -- & $-0.122$ to $0.000$ & $-12$ to $-10$ & $-21\%$ to $-18\%$ \\
    Line 14 at level $\alpha$ & yes & -- & -- & $-0.338$ to $+0.019$ & $0$ & $0\%$ \\
    \midrule
    Phase II removed & no & same & $-2\%$ to $0\%$ & $-0.852$ to $-0.299$ & $0$ & $-7\%$ to $-4\%$ \\
    Line 11 as in the main text & no & same & $-2\%$ to $0\%$ & $-0.383$ to $+0.014$ & $0$ & $-3\%$ to $-2\%$ \\
    Line 13 in descending time order & yes & same & $0\%$ & $-0.052$ to $-0.001$ & $0$ & $+1\%$ \\
    \midrule
    Line 7 removed & yes & same & $+34\%$ to $+69\%$ & $0.000$ & $-2.0$ to $-1.3$ & $+6\%$ to $+23\%$ \\
    Lines 7 and 8 removed & yes & same & $-25\%$ to $-10\%$ & $0.000$ to $+0.011$ & $+0.2$ to $+0.9$ & $-7\%$ to $-3\%$ \\
    \bottomrule
    \end{tabular}
    \caption{One change at a time to ITPD when $N = 20$. $T = 16$; $M \in \{200, 500, 2000\}$; $d \in \{1, 2, 3, 5\}$; 20 graphs per setting with data. 
    }
    \label{tab:ablation}
\end{table}
\begin{table}[!htb]
    \centering
    \scriptsize
    \setlength{\tabcolsep}{2pt}
    \begin{tabular}{@{}c l >{\raggedright\arraybackslash}p{0.3\linewidth} l c c c c c c c@{}}
    \toprule
    & & & & & \multicolumn{5}{c}{Data} & Oracle \\
    \cmidrule(lr){6-10} \cmidrule(lr){11-11}
    Ver. & Line & Change & Status & Exact & F1 & F1$^\ast$ & Tests & $|\mathbf{S}|$ & Set & $|\mathbf{S}|$ \\
    \midrule
    0 &  & Line 10 adds $Z$ with its learned ancestors; Line 9' reads $\mathbf{A}$ & Earlier form & no & 0.741$_{\pm .005}$ & 0.728$_{\pm .004}$ & 1.13 & 110 & 275 & 112 \\
    1 & Line 10 & Adds only $Z$ to $\mathbf{S}$ & Earlier form & no & 0.922$_{\pm .004}$ & 0.864$_{\pm .003}$ & 1.28 & 13 & 30 & 112 \\
    \midrule
    2 & Line 9' & A CI test of $Z$ and $X$ & Earlier draft & no & 0.913$_{\pm .004}$ & 0.859$_{\pm .004}$ & 1.82 & 14 & 33 & 112 \\
    \midrule
    3 & Line 11 & Only $X$ becomes a parent without a CI test & Adopted & yes & 0.926$_{\pm .004}$ & 0.865$_{\pm .003}$ & 1.83 & 14 & 33 & 112 \\
    4 & Line 14 & Level $\alpha_2 = \alpha / (N t - 1)$ & Adopted & yes & 0.996$_{\pm .002}$ & 0.895$_{\pm .003}$ & 1.83 & 14 & 33 & 112 \\
    5 & Lines 8, 10', 10 & Level $\alpha_1$ & Adopted & yes & 0.996$_{\pm .002}$ & 0.946$_{\pm .003}$ & 2.17 & 44 & 91 & 112 \\
    6 & Line 13 & Visits $\mathbf{Z}'$ in ascending time order & Adopted & yes & 0.996$_{\pm .002}$ & 0.960$_{\pm .002}$ & 2.17 & 44 & 91 & 112 \\
    \midrule
    F & Line 9' & From 6: Line 9' deleted & Not adopted & yes & 0.997 & 0.959 & 1.70 & 34 & 71 & 112 \\
    7 & Lines 9', 9 & From 6: Line 9' deleted; Line 9, given $\mathbf{S}$, replaces Line 10' (Algorithm~\ref{alg:itpd}) & Main text & yes & 0.997$_{\pm .001}$ & 0.961$_{\pm .002}$ & 1.38 & 15 & 33 & 3.0 \\
    8 & After Line 10' & From 6: one added test given $\mathbf{S}$ & Not adopted & -- & 0.997$_{\pm .001}$ & 0.960$_{\pm .002}$ & 2.20 & 14 & 31 & 3.0 \\
    \bottomrule
    \end{tabular}
    \caption{\small Exactness and results of each version. Column \textit{Exact} for Versions 6 and 7 and row F 
    is checked on 2{,}359{,}296 enumerated graphs. \textit{Data} columns are means over 20 graphs and the \textit{Oracle} column over 50 graphs. Subscripts of F1 (Versions 0 to 8) are the half-widths of the 95\% intervals. }
    \label{tab:pm_versions}
\end{table}




\end{document}